\pdfoutput=1

\documentclass[10pt,twocolumn,letterpaper]{article}

\usepackage[pagenumbers]{cvpr} %

\usepackage{cuted} %

\usepackage{currfile} %

\usepackage{caption} %

\DeclareMathOperator{\clip}{clip}

\usepackage{dsfont}           %
\newcommand{\ind}{\mathds{1}}

\usepackage{algorithm}
\usepackage{algpseudocode}

\usepackage{multirow}

\usepackage{placeins}

\definecolor{cvprblue}{rgb}{0.21,0.49,0.74}
\usepackage[pagebackref,breaklinks,colorlinks,allcolors=cvprblue]{hyperref}

\def\confName{CVPR}
\def\confYear{2027}

\title{DexDuet: Learning Contact-Preserving Bimanual Dexterity from Human Demonstrations}

\author{Haoran Zhang$^{1}$\quad Haokai Ding$^{1}$\quad Han Zhou$^{1}$\quad Zuxing Lu$^{1}$\quad Heng Zhang$^{2}$\quad Xingxing Zuo$^{1,*}$\\[4pt]
$^{1}$Mohamed bin Zayed University of Artificial Intelligence\qquad $^{2}$Motphys\\[2pt]
{\small $^{*}$Corresponding author}
}

\begin{document}
\raggedbottom
\maketitle
\begin{strip}
  \centering
  \includegraphics[width=\linewidth]{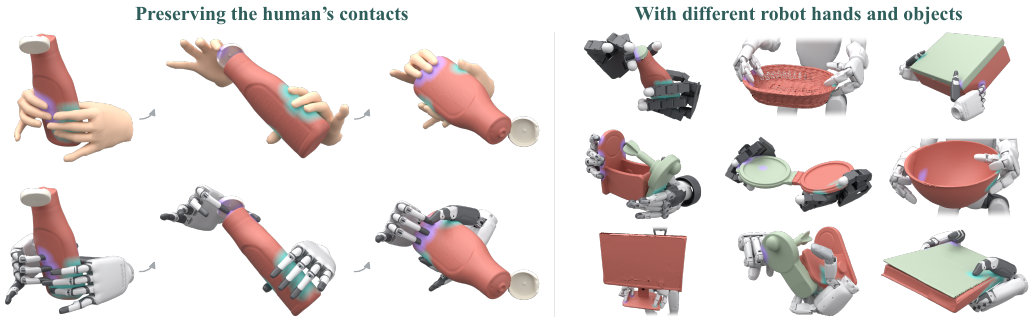}
  \captionsetup{hypcap=false}
  \captionof{figure}{\textbf{DexDuet preserves the human's hand-object contacts.} From one human demonstration of opening a ketchup bottle and handing it over (top left), DexDuet learns a policy for two cooperating robot hands (bottom left) that hold the same parts of the bottle as the human's hands, shown as soft patches for the left hand (teal) and the right hand (violet). DexDuet does the same with other robot hands and objects, including a humanoid with dexterous hands (right).}
  \captionsetup{hypcap=true}
  \label{fig:teaser}
\end{strip}
\begin{abstract}
Human demonstrations provide a rich source of bimanual dexterous skills, yet transferring them to robot hands remains challenging. Existing methods primarily optimize object trajectory tracking, often overlooking how human hands interact with the object. Even with accurate tracking, robots may contact incorrect object parts or slide their fingers, resulting in unstable manipulation.
We present \textbf{DexDuet}, a framework that learns contact-preserving bimanual dexterity from a single human demonstration.
DexDuet employs two interaction rewards to preserve human hand-object configurations and contacts. To enable coordinated manipulation, we factorize the bimanual policy into two weight-sharing per-hand policies that condition on each other's states. We further propose assisted reference-state initialization (A-RSI), an adaptive curriculum that adjusts object assistance and training starts across demonstration segments to overcome difficult manipulation phases.
Across four benchmarks spanning three simulators and five robot embodiments, DexDuet improves object-tracking ADD-AUC by 11--41 points over the corresponding baselines while consistently achieving better contact preservation.
Notably, among 122 paired runs with comparable object-tracking accuracy, DexDuet preserves the human's contacts better in 117 cases.
Code and videos are available on our \href{https://rcl-robotics.github.io/Dexduet.io/}{project page}.
\end{abstract}
    
\section{Introduction}
\label{sec:intro}
People routinely manipulate objects with both hands, such as holding a bottle while opening its lid. Learning such dexterous skills from human demonstrations offers a natural approach to robotic manipulation~\cite{rajeswaran2018dapg,qin2022dexmv,wang2024dexcap,chen2024objdex}.
However, differences between human and robot hands prevent demonstrations from being directly replayed on robots.
Recent methods bridge this gap by retargeting human motions to robot hands and training policies in simulation to track demonstrated object trajectories~\cite{mandi2026dexmachina,li2025maniptrans,zhu2026chord}.
Nevertheless, a fundamental limitation of these methods is their reliance on object trajectory tracking as the primary learning objective.
Even when a robot accurately tracks the demonstrated trajectory, its hands may contact the wrong object parts or slide on the object, resulting in unstable manipulation.
Faithful bimanual manipulation requires not only reproducing \emph{where the object moves}, but also preserving \emph{how the hands interact with it}.

However, achieving contact-preserving bimanual manipulation presents three key challenges.
First, bimanual manipulation requires coordinated yet role-specific control.
Existing methods either control both hands through a unified policy~\cite{mandi2026dexmachina,zhu2026chord} or train separate policies independently~\cite{zhou2024bidexhd}, limiting explicit coordination or shared learning between hands.
Second, preserving human hand-object contacts across different embodiments remains difficult.
Although existing methods incorporate contact-related rewards~\cite{mandi2026dexmachina,zhu2026chord,liang2026contrack}, robot hands can still contact the object differently from human hands, particularly in which hand holds an object part and whether the contact is maintained.
Third, learning complex demonstrations is hindered by difficult manipulation phases.
To make training tractable, prior methods assist the object early in training, with a virtual object controller (VOC), a proportional-derivative (PD) controller that pulls the object toward its demonstrated trajectory~\cite{mandi2026dexmachina,zhu2026chord}, or with relaxed physics~\cite{li2025maniptrans}, and gradually withdraw this assistance.
However, the assistance is typically reduced uniformly across the demonstration, allowing a difficult phase, such as a handover, to stall the entire training process.

To address these challenges, we introduce \textbf{DexDuet}, a framework for learning contact-preserving bimanual dexterity from a single human demonstration (\cref{fig:teaser}).
\textit{To coordinate the two hands}, DexDuet factorizes the bimanual policy into two per-hand policies that share network parameters while observing each other's states.
This allows each hand to perform its own role, adapt to its partner, and benefit from shared learning.
\textit{To preserve human hand-object interactions}, DexDuet introduces two complementary interaction rewards.
The first encourages each robot hand to match the orientation and finger configuration of its human counterpart relative to the object.
The second encourages each hand region, such as the palm or a finger, to contact the object where the corresponding human region does.
Together, these rewards guide the robot to reproduce the demonstrated contacts rather than merely track object motion.
Furthermore, \textit{to facilitate learning through difficult manipulation phases}, we propose assisted reference-state initialization (A-RSI), an adaptive curriculum that adjusts VOC assistance separately for each demonstration segment based on its learning progress.
It also initializes episodes from previously reached states and prioritizes failing segments, enabling the policy to learn complex demonstrations without assistance.
In summary, our contributions are:
\begin{itemize}
  \item We present \textbf{DexDuet}, a framework for learning contact-preserving bimanual dexterity from human demonstrations through factorized, weight-sharing per-hand policies that enable role-specific control and partner-aware coordination.
  \item We introduce two complementary interaction rewards that preserve human hand-object interactions by aligning object-relative hand configurations and region-level contacts across different robot embodiments.
  \item We propose assisted reference-state initialization (A-RSI), an adaptive curriculum that combines segment-specific VOC assistance, initialization from previously reached states, and failure-aware sampling to improve learning through difficult manipulation phases.
  \item We evaluate DexDuet on four benchmarks spanning three simulators and five robot embodiments, demonstrating consistent improvements in object tracking and contact preservation, with additional demonstrations on two real robot systems.
\end{itemize}

\section{Related Work}
\label{sec:related}

\paragraph{Learning dexterous manipulation from human demonstrations.}
Human demonstrations are a natural source of dexterous skills, because people handle everyday objects with both hands skillfully.
Many works therefore collect demonstrations of hand-object interaction.
They capture them in motion-capture and multi-camera studios~\cite{taheri2020grab,chao2021dexycb,fan2023arctic} or with wearable gloves, exoskeletons and headsets~\cite{wang2024dexcap,xu2025dexumi,tao2025dexwild,banerjee2025hot3d}, reconstruct them from monocular or egocentric videos~\cite{qin2022dexmv,fan2024hold,zeng2026stablehand,paliwal2026doasido}, or generate them with models trained on such data~\cite{zeng2026flowhoi}.
These efforts have produced large datasets of hand-object interaction~\cite{liu2024taco,zhan2024oakink2,fu2025gigahands,hoque2026egodex}.
However, these demonstrations cannot be replayed directly on a robot, because robot hands differ from human hands in size, kinematics and actuation.
For humanoid robots and physics-based characters, a common solution to this gap maps the human motion onto the target body and trains a policy in physics simulation to track it~\cite{he2024h2o,liao2025beyondmimic,wang2023physhoi,xu2025intermimic}.
Recent methods apply the same idea to dexterous hands.
They train policies that move the object along its demonstrated trajectory, guided by the retargeted hand motion~\cite{liu2025dextrack,chen2024objdex,li2025maniptrans,lum2025human2sim2robot,mandi2026dexmachina,xu2026demobot,zhu2026chord,liang2026contrack,feng2026regrind,ren2026c2dex}, or optimize the control sequence directly~\cite{liu2024quasisim,pan2025spider}.
These methods are primarily driven by an object-tracking reward.
Some of them also compare the robot's contacts with the human's frame by frame, either by their locations~\cite{lakshmipathy2024kinematic,mandi2026dexmachina,liang2026contrack} or by the forces and torques they can exert on the object~\cite{zhu2026chord}.
These terms act on individual contacts in each frame, and none of them considers whether a contact stays fixed on the object over time.
The robot can therefore still move the object along its trajectory with different contacts, for example with the other hand or with sliding fingers, which makes the manipulation unstable.
DexDuet instead trains each robot hand to preserve the human's contacts, so that it stays on the part that the human's hand holds without sliding.

\paragraph{Bimanual manipulation.}
Two hands that manipulate one object must coordinate, because each performs its own part of the task and the task fails as soon as either one fails.
Imitation-learning policies for two robot arms build this coordination into their architecture, for example with per-arm modules that exchange information or predict a shared future state~\cite{lee2024interact,motoda2025iace,jiang2025decoupled,chen2025statediffusion,xu2025imaginative,grannen2023stabilize,yu2024bikc,xu2026bikcplus}.
In reinforcement learning of bimanual dexterous tasks, coordination has been eased by treating each hand as a separate agent~\cite{chen2022bidexhands}, by giving the two hands complementary roles~\cite{yang2024asymdex}, or by training one hand before both~\cite{zhang2024artigrasp}.
Methods that transfer human demonstrations to two dexterous hands have not used these ideas.
They either drive both hands with one policy, possibly on top of per-hand imitators~\cite{mandi2026dexmachina,zhu2026chord,li2025maniptrans}, or learn an independent policy for each hand~\cite{zhou2024bidexhd}.
With a single policy, the learning of one hand is tied to that of the other, whereas independent policies do not share what one hand learns with the other.
DexDuet lies between these two designs, inspired by cooperative multi-agent reinforcement learning~\cite{yu2022mappo} and by decentralized agents that reason about their partner~\cite{he2025latenttom}.
Each hand has its own policy, which observes the other hand, and the two policies share one network, so that both coordination and experience sharing are built into the policy.

\paragraph{Curricula for dexterous manipulation.}
Learning dexterous manipulation from complex demonstrations is hard, because a policy that starts at the first frame rarely reaches the later phases.
Tracking methods therefore also start episodes at later frames of the demonstration, either from reference states~\cite{zhu2026chord} or from states the policy has already reached~\cite{xu2025intermimic,liang2026contrack,xu2026demobot}, and some sample the start frames more often where the policy fails~\cite{liao2025beyondmimic}.
A second technique assists the object, with a VOC~\cite{mandi2026dexmachina,zhu2026chord} or with relaxed physics~\cite{li2025maniptrans,liu2024quasisim}, and reduces this assistance over training.
However, the assistance is reduced uniformly across the entire demonstration, even though the difficulty usually lies in only a few phases, so that progress on the whole demonstration is limited by its hardest phase.
A-RSI instead sets the assistance separately for each segment of the demonstration, lowering it when the episodes that start there succeed and raising it when they fail, and starts episodes preferentially in segments with high failure rates.
A hard phase thus keeps its assistance and receives more practice, and training on the rest of the demonstration proceeds independently of it.

\section{Method}
\label{sec:method}

\begin{figure*}[t]
  \centering
  \includegraphics[width=\linewidth]{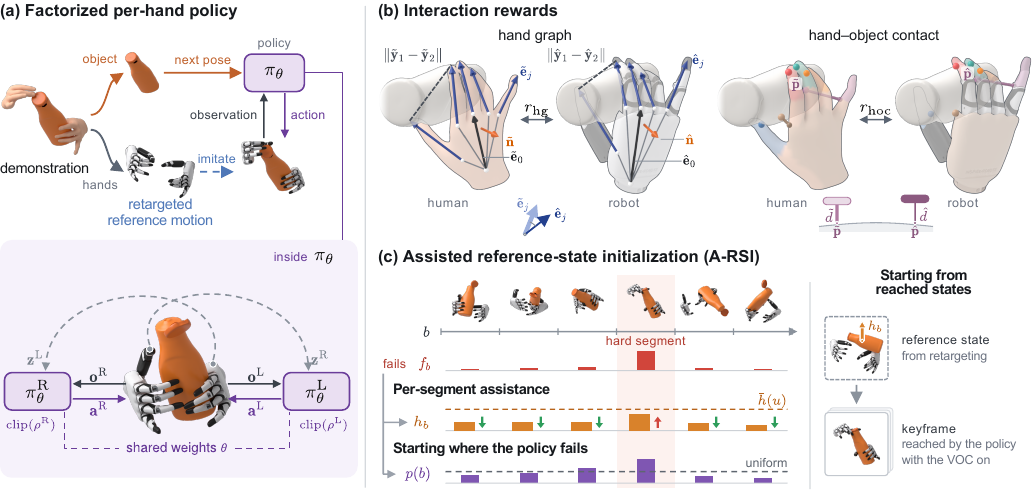}
  \caption{\textbf{Overview of DexDuet.} (a)~A human demonstration provides the object trajectory, which gives the policy $\pi_\theta$ the next object pose, and the hand motion, which is retargeted to a reference motion that the robot hands imitate. Inside $\pi_\theta$, two per-hand policies share one network, and each maps its own observation $\mathbf o$ and the partner observation $\mathbf z$ to an action $\mathbf a$. (b)~Two interaction rewards compare each robot hand with the human hand relative to the object. The hand graph reward compares the palm and finger directions and the fingertip gaps, and the hand-object contact reward compares the distance and the closest point of each hand region to the object. (c)~A-RSI gives each segment $b$ of the demonstration its own assistance level $h_b$, which it lowers (green) or raises (red), never above a ceiling $\bar h(u)$ that decreases over the updates $u$. Episodes start from states that the policy has reached, and their start segment is drawn from $p(b)$, which favors the segments with a high failure rate $f_b$.}
  \label{fig:method}
\end{figure*}

\subsection{Problem Formulation}
\label{sec:method-problem}

Given a human demonstration of a bimanual manipulation, we aim to learn a policy $\pi$ with which two dexterous robot hands reproduce it (\cref{fig:method}).
The demonstration has $T$ frames.
Frame $t$ contains the poses of the two human hands $i\in\{\mathrm L,\mathrm R\}$ and the object state $\mathbf x_t$, which comprises the pose $\mathbf X^m_t\in SE(3)$ of each object part $m$.

Reproducing the demonstration involves two requirements.
First, the object must follow its demonstrated states.
Second, the robot hands must preserve the human's hand-object contacts.
We say that a hand \emph{holds} a part when it touches the part and its contact points do not move on the part.
Contact preservation requires that, whenever a human hand holds a part, the corresponding robot hand holds the same part.
We formalize the two requirements as
\begin{equation}
  \begin{aligned}
    \pi^\star=\arg\min_{\pi}\ &\mathbb E_{\pi}\Big[\textstyle\sum_{t=1}^{T-1}\mathrm{err}_{\text{obj}}(\hat{\mathbf x}_t,\mathbf x_t)\Big]\\
    \text{s.t.}\ \ &\mathbb E_{\pi}\big[\hat\delta_w\big]\le\delta_{\max}\quad\forall\,w\in\mathcal W,
  \end{aligned}
  \label{eq:goal}
\end{equation}
with the object error
\begin{equation}
  \mathrm{err}_{\text{obj}}(\hat{\mathbf x}_t,\mathbf x_t)=\sum_m\frac{1}{|\mathcal S_m|}\sum_{\mathbf s\in\mathcal S_m}\big\lVert(\hat{\mathbf X}^m_t-\mathbf X^m_t)\,\mathbf s\big\rVert,
  \label{eq:obj-err}
\end{equation}
where $\hat{\mathbf x}_t$ is the object state reached by the robot and $\mathcal S_m$ is a set of points on part $m$, expressed in the coordinate frame of the part.
The constraint uses the \emph{contact drift} of a window $w=(i,m,t)$, which measures how far the contact points of hand $i$ on part $m$ at frame $t$ move relative to the part within the next $\Delta$ frames,
\begin{equation}
  \delta_w=\frac{1}{|\mathcal C_w|}\sum_{k\in\mathcal C_w}\big\lVert\mathbf v_{k,t+\Delta}-\mathbf v_{k,t}\big\rVert,
  \label{eq:drift}
\end{equation}
where $\mathcal C_w$ is the set of points of hand $i$ that touch part $m$ at frame $t$ and $\mathbf v_{k,t}$ is the position of point $k$ at frame $t$, expressed in the coordinate frame of the part.
We write $\hat\delta_w$ for the drift of the robot and $\tilde\delta_w$ for that of the human.
A hand holds a part during $w$ if its drift is below a threshold $\delta_{\max}$.
The set $\mathcal W$ contains the \emph{hold windows}, in which the human hand holds the part (\cref{sec:suppl-metrics}).

\subsection{Factorized Per-Hand Policy}
\label{sec:method-policy}

Prior methods control both hands with one unified policy that maps the full observation to the concatenated action of both hands~\cite{mandi2026dexmachina,zhu2026chord}.
Such a policy does not exploit the structure of bimanual manipulation, in which two mirror-symmetric hands act on a shared object and each must act precisely on its own while accounting for the other.
Inspired by decentralized cooperative control, where agents act on their own observations while reasoning about their partner~\cite{he2025latenttom}, we factorize the bimanual policy into two per-hand policies that share one network and observe each other (\cref{fig:method}a).
Each hand $i$ receives a local observation $\mathbf o^i_t$ and a partner observation $\mathbf z^{-i}_t$.
The local observation contains the hand's joint state, wrist pose and keypoints, the object's state and its error to its next demonstrated state, and the hand's previous action.
The keypoints are points on the hand, such as the wrist and the base and tip of each finger.
The partner observation contains the state and previous action of the other hand $-i$ (\cref{sec:suppl-obs}).
The policy over the joint action $\mathbf a_t=(\mathbf a^{\mathrm L}_t,\mathbf a^{\mathrm R}_t)$ is
\begin{equation}
  \pi_\theta(\mathbf a_t\mid\mathbf o_t)=\prod_{i\in\{\mathrm L,\mathrm R\}}\mathcal N\big(\mathbf a^i_t;\ \boldsymbol\mu_\theta(\underbrace{\mathbf o^i_t}_{\text{own hand}},\underbrace{\mathbf z^{-i}_t}_{\text{partner}}),\ \boldsymbol\Sigma^i\big),
  \label{eq:factor}
\end{equation}
where $\mathbf o_t$ collects the observations of both hands, $\mathbf a^i_t$ is the action of hand $i$, which sets its joint position targets, $\boldsymbol\mu_\theta$ is the actor network shared by the two hands, and $\boldsymbol\Sigma^i$ is a learned diagonal covariance (\cref{sec:suppl-arch}).
We denote the policy of hand $i$, which is the corresponding factor of \cref{eq:factor}, by $\pi^i_\theta$.

\paragraph{Per-hand clipping.}
We train the policy with PPO~\cite{schulman2017ppo}, which keeps each update small by clipping the probability ratio between the updated and the previous policy.
For a unified policy, this is the ratio of the joint action of both hands, so a sample on which the policy of one hand has changed too much is clipped for both hands and no longer updates the other hand.
We instead clip the ratio of each hand separately (\cref{fig:method}a) and maximize
\begin{equation}
  \begin{gathered}
    \mathbb E_t\Big[\textstyle\sum_{i}\min\big(\rho^i_t A^i_t,\ \clip_\epsilon(\rho^i_t)\,A^i_t\big)\Big],\\
    \rho^i_t=\pi^i_\theta(\mathbf a^i_t\mid\mathbf o_t)\,/\,\pi^i_{\theta_{\text{old}}}(\mathbf a^i_t\mid\mathbf o_t),
  \end{gathered}
  \label{eq:ppo}
\end{equation}
where $\rho^i_t$ is the probability ratio of hand $i$, $\theta_{\text{old}}$ are the parameters before the update, $\clip_\epsilon$ clips a ratio to $[1-\epsilon,1+\epsilon]$, and $A^i_t$ is the advantage of hand $i$, estimated by its own value head of a centralized critic (\cref{sec:suppl-ppo}).

\subsection{Interaction Rewards}
\label{sec:method-reward}

We build on the rewards commonly used in prior work~\cite{mandi2026dexmachina,zhu2026chord}, which track the object and imitate a \emph{reference motion} obtained by retargeting the human hands to the robot hands, and denote their sum by $r_{\text{base}}$ (\cref{sec:suppl-reward}).
To preserve the human's hand-object contacts, we add two dense terms that compare each robot hand with the human hand itself, relative to the object (\cref{fig:method}b),
\begin{equation}
  r_t=r_{\text{base}}+\lambda_{\text{hg}}\,r_{\text{hg}}+\lambda_{\text{hoc}}\,r_{\text{hoc}},
  \label{eq:reward}
\end{equation}
with weights $\lambda_{\text{hg}}$ and $\lambda_{\text{hoc}}$ (\cref{tab:suppl-reward}).
They are inspired by the contact-graph rewards for physics-based human-object interaction~\cite{wang2023physhoi}.

\paragraph{Hand graph reward.}
This term aligns the palm and fingers of each robot hand with those of the human hand, in their orientation and opening around the object.
We build a graph on the keypoints of each hand.
Its edges $\mathcal E$ are unit vectors and consist of the palm normal $\mathbf n$, the palm direction $\mathbf e_0$ from the wrist to the base of the middle finger, and the direction $\mathbf e_j$ of each finger $j$ from its base to its tip.
The edges of the human hand are taken relative to the demonstrated object, and those of the robot hand relative to the reached object.
Two errors compare the robot hand with the human hand,
\begin{equation}
  \begin{gathered}
    \mathrm{err}_{\text{dir}}=\sum_{\mathbf e\in\mathcal E}\omega_{\mathbf e}\,\big(1-\hat{\mathbf e}^{\top}\tilde{\mathbf e}\big),\\
    \mathrm{err}_{\text{gap}}=\frac1{|\mathcal P|}\!\sum_{(j,j')\in\mathcal P}\!\Big|\lVert\hat{\mathbf y}_j-\hat{\mathbf y}_{j'}\rVert-\kappa\,\lVert\tilde{\mathbf y}_j-\tilde{\mathbf y}_{j'}\rVert\Big|,
  \end{gathered}
  \label{eq:handgraph-err}
\end{equation}
The direction error compares how the palm and each finger are oriented.
In it, $\hat{\mathbf e}$ and $\tilde{\mathbf e}$ denote the same edge of the robot hand and of the human hand, and the weights $\omega_{\mathbf e}$ sum to one.
The gap error compares how far the fingers are spread.
In it, $\hat{\mathbf y}_j$ and $\tilde{\mathbf y}_j$ are the positions of the tip of finger $j$, $\mathcal P$ contains the pairs of the thumb with each other finger and of neighboring fingers, and $\kappa$ scales the human hand to the size of the robot hand (\cref{sec:suppl-reward}).
With scales $\beta_{\text{dir}}$ and $\beta_{\text{gap}}$, the reward combines the two errors of each hand $i$ and averages over the hands,
\begin{equation}
  r_{\text{hg}}=\tfrac12\textstyle\sum_i e^{-\beta_{\text{dir}}\,\mathrm{err}^i_{\text{dir}}-\beta_{\text{gap}}\,\mathrm{err}^i_{\text{gap}}}.
  \label{eq:handgraph}
\end{equation}

\paragraph{Hand-object contact reward.}
This term encourages each region of the robot hand to approach and contact the object where the corresponding region of the human hand does.
We divide each hand into the palm, the thumb and each of the other fingers, and index these regions by $g$.
For a hand with fewer fingers than the human's, both rewards use only the fingers that the retargeting matches (\cref{fig:suppl-fingers}).
For each region $g$ and object part $m$, we compute the distance $d_{g,m}$ from the region to the part and the point $\mathbf p_{g,m}$ on the part that is closest to the region.
We compute both quantities for the human hand, $\tilde d_{g,m}$ and $\tilde{\mathbf p}_{g,m}$, and for the robot hand, $\hat d_{g,m}$ and $\hat{\mathbf p}_{g,m}$ (\cref{sec:suppl-reward}).
For each hand, the reward compares the robot with the human region by region,
\begin{equation}
  \begin{aligned}
    \mathrm{err}_{g,m}={}&\big|\min(\hat d_{g,m},d_{\max})-\min(\tilde d_{g,m},d_{\max})\big|\\
    &+\ind[\tilde d_{g,m}<d_{\max}]\,\lVert\hat{\mathbf p}_{g,m}-\tilde{\mathbf p}_{g,m}\rVert,\\
    r^i_{\text{hoc}}={}&\textstyle\sum_{g,m}\omega_{g,m}\,e^{-\beta_{\text{hoc}}\max(0,\,\mathrm{err}_{g,m}-d_{\text{tol}})},
  \end{aligned}
  \label{eq:hoc}
\end{equation}
where $d_{\max}$ caps the distances, $d_{\text{tol}}$ is a tolerance, $\beta_{\text{hoc}}$ is a scale, and the weights $\omega_{g,m}$ sum to one and emphasize the pairs of a region and a part that are in or near contact (\cref{sec:suppl-reward}).
The hand index $i$ is omitted on the right-hand sides, and $r_{\text{hoc}}$ is the mean of $r^i_{\text{hoc}}$ over the two hands.
The first term of the error compares how close each region is to the part, and the cap makes all far regions count as equally far.
The second term compares where the region touches the part and applies only where the human region is close to the part.
The reward thus guides each region toward the part before contact is made, and it penalizes contacts that the human does not make.

\subsection{Assisted Reference-State Initialization}
\label{sec:method-curriculum}

Learning a complex bimanual demonstration is a hard exploration problem, because episodes rarely progress beyond a difficult phase in the middle, such as a handover between the hands.
In motion imitation, reference-state initialization (RSI)~\cite{peng2018deepmimic} addresses this problem by starting episodes from states sampled uniformly along the reference motion, so that every phase is practiced.
In dexterous retargeting, a \emph{reference state} consists of the hand configurations of the reference motion and the demonstrated object state of a frame.
RSI then faces two problems.
First, a reference state is usually not physically valid, because the retargeted fingers often leave a gap to the object or penetrate it.
Without a valid grasp, an object that starts in the air falls or is pushed away before the policy can act.
Second, uniform sampling spends most episodes on phases that the policy has already solved instead of on the difficult one.
\emph{Assisted reference-state initialization} (A-RSI, \cref{fig:method}c and \cref{fig:arsi}) addresses the first problem by assisting the object, which makes reference states usable, and by starting episodes from states that the policy has reached.
In addition, it addresses the second problem by starting episodes more often where the policy fails.

\paragraph{Per-segment assistance.}
To make reference states usable, we assist the object with the VOC~\cite{mandi2026dexmachina}, which pulls the object toward its next demonstrated state $\mathbf x_{t+1}$ with a strength set by its gains.
Prior work~\cite{mandi2026dexmachina,zhu2026chord} uses one gain level for the whole demonstration and lowers it only once the policy succeeds over all of it, so a single hard phase stalls the curriculum.
We instead divide the demonstration into $B$ segments of equal length, indexed by $b$, and give each segment its own assistance level $h_b$.
This level sets the VOC gains of every episode that starts in segment $b$, for the whole episode.
An episode succeeds if it reaches the end of the demonstration and fails if it is terminated early because its tracking reward falls below a threshold.
Each $h_b$ starts at a maximum $h_{\max}$ and is updated during training according to the success rate $s_b$ of the recent episodes that started in segment $b$,
\begin{equation}
  h_b\leftarrow
  \begin{cases}
    \alpha\,h_b & \text{if } s_b\ge s_{\text{hi}},\\
    h_b/\alpha & \text{if } s_b<s_{\text{lo}} \text{ and } h_b/\alpha\le\bar h(u),\\
    h_b & \text{otherwise,}
  \end{cases}
  \label{eq:help}
\end{equation}
where $\alpha<1$ is a decay factor, $s_{\text{lo}}<s_{\text{hi}}$ are thresholds, and $\bar h(u)$ is a ceiling on all levels that decreases with the index $u$ of the PPO update, from $h_{\max}$ to zero.
Assistance is thus withdrawn from the segments that the policy has mastered and restored where it fails.
The ceiling guarantees that every level reaches zero, so the final policy runs without assistance.
Training stops once every segment has been unassisted for a fixed number of updates, or at an update cap (\cref{alg:curriculum,sec:suppl-curriculum}).

\paragraph{Starting from reached states.}
Under full assistance, an episode can start from a reference state, because the VOC keeps the object on its trajectory even though the hands do not yet hold it.
As the assistance of a segment is lowered, this is no longer reliable, and the object may be lost at the start of the episode.
A-RSI therefore starts episodes from \emph{keyframes}, which are states that the policy itself has reached and which are therefore physically valid.
During training, a state is stored as a keyframe of its segment if its episode subsequently reaches the end of the demonstration, and the episodes that start in this segment are then initialized from its keyframes instead of reference states.
A keyframe is used only while the assistance of its segment is at least as high as the level under which it was reached (\cref{sec:suppl-curriculum}).

\paragraph{Starting where the policy fails.}
Inspired by the adaptive sampling of BeyondMimic~\cite{liao2025beyondmimic}, A-RSI starts episodes more often in the segments where the policy fails.
An episode starts in segment $b$ with probability
\begin{equation}
  \begin{aligned}
    P_{\text{start}}(b)&=\lambda_0\,\ind[b=0]+(1-\lambda_0)\,p(b),\\
    p(b)&=\frac{\lambda_{\text{u}}}{B}+(1-\lambda_{\text{u}})\,\frac{F_b}{F_0+\dots+F_{B-1}},\\
    F_b&=\textstyle\sum_{k=0}^{B-1-b}\gamma_{\text{seg}}^{\,k}\,f_{b+k},
  \end{aligned}
  \label{eq:startdist}
\end{equation}
where the segments are indexed from $0$ to $B-1$ and $F_b$ is the failure score of segment $b$.
The first term of $P_{\text{start}}$ accounts for the episodes that start at the first frame, with probability $\lambda_0$, and run through the whole demonstration, as the final policy does.
The remaining episodes draw their start segment from $p(b)$, which mixes a uniform distribution over the $B$ segments, with weight $\lambda_{\text{u}}$, and a distribution proportional to the failure scores $F_b$.
Here $f_b$ is the failure rate of the recent episodes that started in segment $b$, and $F_b$ adds to it the failure rates of the segments after $b$ with a discount $\gamma_{\text{seg}}$, so that the segments leading up to a failing phase are also practiced.

\section{Experiments}
\label{sec:exp}

\begin{table*}[t]
  \caption{\textbf{Comparison with prior methods.} All metrics are in \% except MPPE (cm). Each entry is the mean over demonstrations $\pm$ the std over seeds, with five seeds on the DexMachina and SPIDER suites and three on the ManipTrans suite. On V2D, each method is trained with one seed, and each entry is the mean over the 30 episodes of four runs of the challenge's official evaluator $\pm$ the std over these four runs. Numbers are not comparable across benchmarks.}
  \label{tab:main}
  \centering
  \footnotesize
  \setlength{\tabcolsep}{6pt}
  \newcommand{\stdv}[1]{{\scriptsize$\pm$#1}}
  \begin{tabular}{@{}llccccc@{}}
    \toprule
    Robot & Method & ADD-AUC$\uparrow$ & SP-SR$\uparrow$ & MP-SR$\uparrow$ & MPPE$\downarrow$ & Contact-AUC$\uparrow$ \\
    \midrule
    \multicolumn{7}{@{}l}{\textit{DexMachina suite}~\cite{mandi2026dexmachina}} \\
    \multirow{2}{*}{Inspire} & DexMachina & 52.5\stdv{10.2} & 53.7\stdv{11.8} & 53.7\stdv{11.8} & 40.7\stdv{15.6}& 33.6\stdv{7.8} \\
                             & DexDuet    & \textbf{88.7}\stdv{2.3} & \textbf{94.3}\stdv{7.8} & \textbf{94.3}\stdv{7.8} & \textbf{1.50}\stdv{0.22} & \textbf{79.6}\stdv{2.6} \\
    \addlinespace[2pt]
    \multirow{2}{*}{Allegro} & DexMachina & 65.8\stdv{14.4} & 67.9\stdv{18.4} & 67.9\stdv{18.4} & 23.4\stdv{15.4}& 35.8\stdv{8.3} \\
                             & DexDuet    & \textbf{94.8}\stdv{0.9} & \textbf{100.0}\stdv{0.0} & \textbf{100.0}\stdv{0.0} & \textbf{1.02}\stdv{0.09} & \textbf{68.0}\stdv{1.9} \\
    \addlinespace[2pt]
    \multirow{2}{*}{XHand}   & DexMachina & 60.7\stdv{5.0} & 62.3\stdv{7.7} & 62.2\stdv{7.7} & 28.7\stdv{6.1}& 38.2\stdv{4.4} \\
                             & DexDuet    & \textbf{93.6}\stdv{2.1} & \textbf{100.0}\stdv{0.0} & \textbf{97.1}\stdv{6.4} & \textbf{1.09}\stdv{0.19} & \textbf{79.5}\stdv{3.5} \\
    \addlinespace[2pt]
    \multirow{2}{*}{Schunk}  & DexMachina & 73.7\stdv{8.7} & 79.2\stdv{16.2} & 76.4\stdv{16.1} & 12.1\stdv{11.6}& 45.2\stdv{6.5} \\
                             & DexDuet    & \textbf{94.1}\stdv{0.3} & \textbf{100.0}\stdv{0.0} & \textbf{100.0}\stdv{0.0} & \textbf{1.06}\stdv{0.02} & \textbf{80.4}\stdv{2.2} \\
    \midrule
    \multicolumn{7}{@{}l}{\textit{SPIDER suite}~\cite{pan2025spider}} \\
    \multirow{2}{*}{Inspire} & SPIDER & 47.5\stdv{2.9} & 34.3\stdv{7.8} & 31.4\stdv{6.4} & 44.33\stdv{3.40} & 36.5\stdv{2.2} \\
                             & DexDuet    & \textbf{88.5}\stdv{4.5} & \textbf{97.1}\stdv{6.4} & \textbf{94.3}\stdv{7.8} & \textbf{3.61}\stdv{5.39} & \textbf{76.9}\stdv{4.8} \\
    \midrule
    \multicolumn{7}{@{}l}{\textit{ManipTrans suite}~\cite{li2025maniptrans}} \\
    \multirow{2}{*}{Inspire} & ManipTrans & 87.0\stdv{4.1} & 87.1\stdv{4.2} & 84.9\stdv{4.2} & 4.68\stdv{1.31} & 72.6\stdv{5.7} \\
                             & DexDuet    & \textbf{98.4}\stdv{0.4} & \textbf{99.9}\stdv{0.1} & \textbf{99.9}\stdv{0.1} & \textbf{0.44}\stdv{0.02} & \textbf{82.7}\stdv{1.1} \\
    \midrule
    \multicolumn{7}{@{}l}{\textit{V2D challenge}~\cite{nvidia2026v2d}} \\
    \multirow{3}{*}{G1 + Dex3} & CHORD (PPO)~\cite{zhu2026chord} & 39.8\stdv{0.3} & 53.3\stdv{0.0} & 15.8\stdv{5.0} & 14.85\stdv{1.29} & 16.0\stdv{1.1} \\
                               & CHORD (FlashSAC)~\cite{zhu2026chord} & 32.0\stdv{0.3} & 35.0\stdv{3.3} & 12.5\stdv{1.7} & 37.77\stdv{2.07} & 13.0\stdv{0.4} \\
                               & DexDuet    & \textbf{77.4}\stdv{0.9} & \textbf{85.0}\stdv{1.9} & \textbf{75.8}\stdv{3.2} & \textbf{12.19}\stdv{0.93} & \textbf{72.6}\stdv{0.8} \\
    \bottomrule
  \end{tabular}
\end{table*}

Our experiments address four questions:
\begin{enumerate}[label=\textbf{Q\arabic*},leftmargin=*,noitemsep,topsep=2pt]
  \item Does DexDuet track the object better than the compared baselines for different benchmarks?
  \item Does it preserve the human's hand-object contacts?
  \item Can the learned motions be executed on real robot hands?
  \item What do the per-hand policy, A-RSI and the interaction rewards contribute, and at what training cost?
\end{enumerate}

\subsection{Experimental Setup}
\label{sec:exp-setup}

\paragraph{Benchmarks and baselines.}
We evaluate on three dexterous manipulation benchmarks and on a humanoid manipulation challenge, each in the simulator of its baseline.
The DexMachina~\cite{mandi2026dexmachina} and SPIDER~\cite{pan2025spider} suites each have their own seven ARCTIC~\cite{fan2023arctic} demonstrations in Genesis~\cite{genesis2024}.
We evaluate DexMachina's released checkpoints for all four hands and rerun SPIDER's official code, which optimizes an open-loop control sequence instead of a policy.
The ManipTrans suite~\cite{li2025maniptrans} has ten OakInk2~\cite{zhan2024oakink2} demonstrations in Isaac Gym~\cite{makoviychuk2021isaacgym}, and we retrain ManipTrans with its official protocol.
The V2D challenge~\cite{nvidia2026v2d} has 30 humanoid episodes in MuJoCo Warp~\cite{todorov2012mujoco,mujocowarp2025}, and we retrain its baseline CHORD~\cite{zhu2026chord} with the challenge's official PPO and FlashSAC~\cite{kim2026flashsac} recipes at the same budget of environment steps as DexDuet (\cref{sec:suppl-bench}).

\paragraph{Metrics.}
We report the following four object-tracking metrics: ADD-AUC~\cite{hinterstoisser2012add,xiang2018posecnn,mandi2026dexmachina}, the mean per-point error of the object (MPPE), and the success rates SP-SR and MP-SR.
SP-SR uses the thresholds of SPIDER, and MP-SR those of ManipTrans.
These metrics evaluate only the first requirement of \cref{sec:method-problem}, how closely the object follows its demonstrated states.
They do not reveal whether the robot hands preserve the human's hand-object contacts.
We therefore introduce a further metric, Contact-AUC, which measures how well the constraint of \cref{eq:goal} is met.
For every hold window $w\in\mathcal W$, it takes the contact drift $\hat\delta_w$ of the corresponding robot hand and, as ADD-AUC does for the object's error, scores the share of thresholds $\tau$ between $\tau_{\min}$ and $\tau_{\max}$ that the drift stays below,
\begin{equation}
  \begin{aligned}
    \text{Contact-AUC}&=\frac{1}{|\mathcal W|}\sum_{w\in\mathcal W}\mathrm{AUC}_w,\\
    \mathrm{AUC}_w&=\tfrac{1}{\tau_{\max}-\tau_{\min}}\textstyle\int_{\tau_{\min}}^{\tau_{\max}}\ind\big[\hat\delta_w<\tau\big]\,d\tau,
  \end{aligned}
  \label{eq:contact-auc}
\end{equation}
averaged over rollouts.
A window scores 0 if the robot hand does not touch the part or if the object is lost.
The window length and all thresholds are given in \cref{sec:suppl-metrics}.

\subsection{Comparison with Prior Work}
\label{sec:results}

\begin{figure*}[t]
  \centering
  \includegraphics[width=\linewidth]{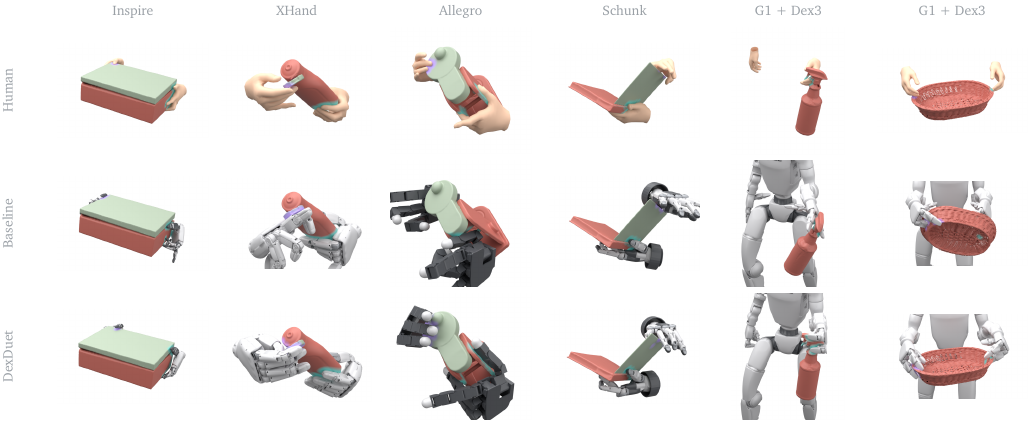}
  \caption{\textbf{Qualitative comparison with the baselines.} The robot is named above each column. The first four columns are Box-200, Ketchup-100, Mixer-170 and Notebook-300 from the DexMachina suite. Each pairs a DexMachina~\cite{mandi2026dexmachina} checkpoint with a DexDuet run whose ADD-AUC differs by at most 3 points. The last two columns are the spray bottle and basket episodes of V2D, where the baseline is CHORD~\cite{zhu2026chord}. Soft patches mark where the left (teal) and right (violet) hand touch the object.}
  \label{fig:qualitative}
\end{figure*}

\begin{table}[t]
  \caption{\textbf{Contact preservation at equal tracking and holding after the last frame} (DexMachina suite). Equal tracking: pairs of a DexDuet and a baseline run on the same hand and demonstration whose ADD-AUCs differ by at most 3 points. Higher: pairs in which DexDuet has the higher Contact-AUC, of all pairs. Gain: median difference in Contact-AUC, in points. Hold rate: share of rollouts that keep the object within 5\,cm of its final pose for 3\,s after the last frame, with every reference frozen, in \%, mean $\pm$ std over training seeds. Details are in \cref{sec:suppl-matched,sec:suppl-hold}.}
  \label{tab:hold}
  \centering
  \footnotesize
  \newcommand{\stdv}[1]{{\scriptsize$\pm$#1}}
  \begin{tabular}{@{}lcccc@{}}
    \toprule
    & \multicolumn{2}{c}{Equal tracking} & \multicolumn{2}{c}{Hold rate} \\
    \cmidrule(lr){2-3} \cmidrule(l){4-5}
    Hand & Higher & Gain & DexMachina & DexDuet \\
    \midrule
    Inspire & 28 / 28   & $+22.4$ & 17.0\stdv{11.8} & \textbf{37.1}\stdv{16.3} \\
    Allegro & 41 / 42   & $+20.0$ & 14.1\stdv{10.0} & \textbf{22.9}\stdv{12.8} \\
    XHand   & 19 / 20   & $+12.4$ & 5.7\stdv{7.7}   & \textbf{25.7}\stdv{15.6} \\
    Schunk  & 29 / 32   & $+16.8$ & 25.5\stdv{11.8} & \textbf{42.9}\stdv{10.1} \\
    \midrule
    All     & 117 / 122 & $+17.9$ & 15.6\stdv{3.2}  & \textbf{32.1}\stdv{7.1} \\
    \bottomrule
  \end{tabular}
\end{table}

\begin{figure*}[t]
  \centering
  \includegraphics[width=\linewidth]{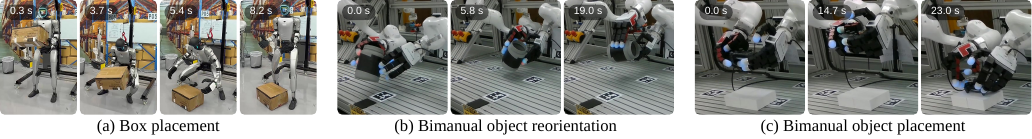}
  \caption{\textbf{Real-world demonstrations.} DexDuet's simulated rollouts replayed on real robots. (a) A Unitree G1 with Inspire hands places a box on the floor. (b) Two Franka arms with Allegro hands reorient a bottle and (c) place a gamepad on a block.}
  \label{fig:real}
\end{figure*}

\paragraph{Q1: Object tracking.}
DexDuet tracks the object better than the baseline of every benchmark and with every robot hand (\cref{tab:main}).
The gain is largest on Ketchup-300 with Inspire hands (4.9 to 86.9, \cref{tab:suppl-dexmachina}), and on the V2D humanoid DexDuet nearly doubles the retrained CHORD's ADD-AUC (\cref{fig:qualitative}, last two columns).

\paragraph{Q2: Contact preservation.}
DexDuet reaches a higher Contact-AUC than the baseline of every benchmark (\cref{tab:main}).
This advantage remains where a baseline tracks the object equally well.
Among the pairs of a DexDuet and a DexMachina run with ADD-AUCs within 3 points, DexDuet has the higher Contact-AUC in 117 of 122, and in at least 90\% of the pairs of each hand (\cref{tab:hold}).
In these pairs, the baseline's hand often leaves the part that the human's hand holds, or its fingers slide on it (\cref{fig:qualitative}, first four columns, and \cref{fig:suppl-matched}).
In a perceptual study on 24 pairs of runs that track equally well and differ by at least 10 points of Contact-AUC, eleven raters judged the run with the higher Contact-AUC to handle the object more like the human in 251 of 264 judgments (\cref{sec:suppl-perceptual}).

DexDuet also holds the object more often after the demonstration ends, with every hand (\cref{tab:hold}).
The baseline ends the demonstration more than 3\,cm from the final pose in 62 of its 140 runs, against 11 for DexDuet (\cref{sec:suppl-hold}).

\subsection{Real-World Demonstrations}
\label{sec:real}

\paragraph{Q3: Real robot hands.}
We replay DexDuet's simulated rollouts open-loop on two real setups (\cref{fig:real}).
On two Franka arms with Allegro hands, both hands hold a bottle above the table and reorient it, following a demonstration from OakInk2~\cite{zhan2024oakink2}, and place a gamepad, held by both hands, on a block.
On a Unitree G1 humanoid with Inspire hands, the robot lowers a box to the floor, puts down a bag and pushes an office chair while walking, following human demonstrations reconstructed from video with Do as I Do~\cite{paliwal2026doasido}.
Its locomotion is handled by a whole-body controller in the style of SONIC~\cite{luo2025sonic}, deployed with RoboJuDo~\cite{zhuang2025robojudo}.
\Cref{sec:suppl-real} shows every replay frame by frame, with videos in the supplementary material.

\subsection{Ablation Studies}
\label{sec:ablation}

\paragraph{Q4: Contribution of each component.}
Each component prevents a different failure (\cref{tab:ablation,sec:suppl-ablation}).
Without the per-hand policy (e), tracking is less precise in 20 of the 21 runs paired by demonstration and seed.
Replacing A-RSI with a single-level VOC curriculum (g) lowers Contact-AUC by 6.3 points and requires 20\% more updates until training stops (\cref{fig:assistance,tab:suppl-ablation}).
Removing either interaction reward (b, c) lowers Contact-AUC much more than ADD-AUC, and most of this loss lies on windows in which the object stays on track, so it comes from the contacts (\cref{tab:suppl-decomp}).
Without both rewards (d), six of the 21 runs fail.

\begin{table}[t]
  \caption{\textbf{Ablations} on the DexMachina suite (Inspire, seven demonstrations, three seeds): ADD-AUC and Contact-AUC in \%, mean $\pm$ std over seeds, and mean PPO updates until training stops.}
  \label{tab:ablation}
  \centering
  \footnotesize
  \setlength{\tabcolsep}{3.5pt}
  \newcommand{\stdv}[1]{{\tiny$\pm$#1}}
  \begin{tabular}{@{}lccccccc@{}}
    \toprule
    & Per-hand & A-RSI & $r_{\text{hg}}$ & $r_{\text{hoc}}$ & ADD$\uparrow$ & Contact$\uparrow$ & Updates$\downarrow$ \\
    \midrule
    (a) & \checkmark & \checkmark & \checkmark & \checkmark & \textbf{89.7}\stdv{2.1} & \textbf{80.5}\stdv{3.2} & \textbf{5,524} \\
    (b) & \checkmark & \checkmark &            & \checkmark & 87.5\stdv{1.1} & 71.5\stdv{1.6} & 6,721 \\
    (c) & \checkmark & \checkmark & \checkmark &            & 86.4\stdv{3.9} & 65.4\stdv{4.5} & 5,836 \\
    (d) & \checkmark & \checkmark &            &            & 76.7\stdv{8.8} & 49.0\stdv{5.1} & 6,745 \\
    (e) &            & \checkmark & \checkmark & \checkmark & 74.1\stdv{1.7} & 62.1\stdv{2.6} & 6,255 \\
    (f) & \checkmark &            &            &            & 79.4\stdv{5.9} & 47.6\stdv{4.5} & 6,845 \\
    (g) & \checkmark &            & \checkmark & \checkmark & 88.4\stdv{1.6} & 74.2\stdv{3.2} & 6,608 \\
    \bottomrule
  \end{tabular}
\end{table}

\section{Conclusion}
\label{sec:conclusion}

Prior methods that learn bimanual dexterity from human demonstrations mainly reward object tracking, and their policies often track the object without preserving the human's hand-object contacts.
We presented DexDuet, which preserves these contacts with two interaction rewards that compare each robot hand with the human hand relative to the object.
To track complex demonstrations, DexDuet also coordinates the two hands with per-hand policies and learns the hard phases with its A-RSI curriculum.
Across four benchmarks and five robot embodiments, DexDuet tracks the object and preserves the human's contacts better than prior methods, and its motions are executed on two real robot systems.

{
    \small
    \bibliographystyle{ieeenat_fullname}
    \bibliography{reference}
}

\backrefsetup{disable}
\clearpage
\raggedbottom
\makeatletter\setlength{\@fptop}{0pt}\setlength{\@dblfptop}{0pt}\setlength{\@fpsep}{18pt}\setlength{\@dblfpsep}{18pt}\setlength{\@fpbot}{0pt plus 1fil}\setlength{\@dblfpbot}{0pt plus 1fil}\makeatother
\maketitlesupplementary
\setcounter{section}{0}\renewcommand{\thesection}{\Alph{section}}\def\theHsection{S\Alph{section}}
\setcounter{figure}{0}\renewcommand{\thefigure}{S\arabic{figure}}\def\theHfigure{S\arabic{figure}}
\setcounter{table}{0}\renewcommand{\thetable}{S\arabic{table}}\def\theHtable{S\arabic{table}}
\setcounter{equation}{0}\renewcommand{\theequation}{S\arabic{equation}}\def\theHequation{S\arabic{equation}}
\renewcommand{\thealgorithm}{S\arabic{algorithm}}

\section{Additional Method Details}
\label{sec:suppl-method}

Unless stated otherwise, the settings of this section apply to two Inspire hands on ARCTIC demonstrations in Genesis.
The Allegro, XHand and Schunk hands keep the curriculum of \cref{tab:suppl-curriculum,tab:suppl-clips} and the weights $\lambda_{\text{hg}}$ and $\lambda_{\text{hoc}}$, and \cref{tab:suppl-hands} lists the settings that depend on the hand.
Their PD gains are those of DexMachina's configuration of each hand, and every reward term that averages over keypoints, joints, links or action entries averages over those of the hand.
On the demonstrations of group A (\cref{tab:suppl-reward}), their reward weights are those of DexMachina's released runs with the same hand, except that our Schunk runs keep the Inspire value $\lambda_{\text{bc}}=0.3$ instead of 0.2; no weight was tuned per hand.

\subsection{Observations and Actions}
\label{sec:suppl-obs}

\paragraph{Object state.}
The ARCTIC objects consist of two rigid parts connected by a revolute joint.
As in ARCTIC~\cite{fan2023arctic} and DexMachina~\cite{mandi2026dexmachina}, we parametrize their state $\mathbf x_t=(\mathbf p_t,\mathbf R_t,\phi_t)$ by the position $\mathbf p_t\in\mathbb R^3$ and orientation $\mathbf R_t\in SO(3)$ of the base part and the joint angle $\phi_t$.

\paragraph{Hands.}
Each hand sits on a floating wrist with three prismatic and three revolute joints, and its action has one dimension for each wrist joint and each independent finger joint.
The Inspire hand has 12 finger joints, 6 of which are mimic joints that copy a parent joint through a fixed ratio.
Its configuration $\mathbf q^i\in\mathbb R^{18}$ therefore has 12 independent DoFs, and each hand has a 12-dimensional action, for 24 in total.
The Allegro, XHand and Schunk hands have 16, 12 and 20 finger joints, of which only the Schunk hand has mimic joints (11), so their actions have 22, 18 and 15 dimensions per hand (\cref{tab:suppl-hands}).
On the V2D challenge, the Dex3 hands of the G1 humanoid are positioned by its arms (\cref{sec:suppl-locomanip}).

\paragraph{Observations.}
\Cref{tab:suppl-obs} lists the observation blocks.
The critic is used only during training.
It receives every block in its 499-dimensional input, including the contact forces and hand-object distances that the actor does not observe.
The actor of hand $i$ receives a 157-dimensional local observation $\mathbf o^i_t$, which holds its own hand, the object and phase and its own previous action, and a 56-dimensional partner observation $\mathbf z^{-i}_t$, which holds the partner's state (43), its previous action (12) and a constant flag (\cref{eq:factor}).
All observations are clipped to $[-5,5]$ without further normalization.
The previous-action block holds the action $\mathbf a^i_{t-2}$ applied two control steps earlier, and is zero at the start of an episode.
The only explicit reference in the actor's input is the object's error to its next demonstrated state.
The actor observes no further future reference.

\begin{table*}[!tp]
  \caption{\textbf{Observation blocks.} Dimensions are per hand unless the block covers both hands.}
  \label{tab:suppl-obs}
  \centering
  \footnotesize
  \setlength{\tabcolsep}{4pt}
  \begin{tabular}{@{}lrp{0.45\linewidth}cc@{}}
    \toprule
    Block & Dim. & Content & Actor $i$ & Critic \\
    \midrule
    Own hand (in $\mathbf o^i_t$) & 115 & PD target minus joint position (18); joint position normalized to $[-1,1]$ by the limits (18); joint velocity $\times0.1$ (18); world positions of the 18 keypoints (54); wrist position and quaternion (7) & own hand & both hands \\
    Object and phase (in $\mathbf o^i_t$) & 30 & poses of the two parts (14); joint angle (1); error to the next demonstrated state $\mathbf x_{t+1}-\hat{\mathbf x}_t$ (8, quaternions subtracted componentwise); base angular velocity $\times0.25$ (3); base linear velocity $\times2$ (3); phase $2t/T-1$ (1) & yes & yes \\
    Previous action (in $\mathbf o^i_t$, $\mathbf z^{-i}_t$) & 12 & $\mathbf a^i_{t-2}$ & own; partner's via $M_\theta$ & both \\
    Partner state (in $\mathbf z^{-i}_t$) & 43 & partner wrist position and orientation in the object frame (3+4); normalized joint positions (18); joint velocities $\times0.1$ (18) & via $M_\theta$ & yes \\
    Keypoint-to-object distance & 72 & for each hand, keypoint and part, the minimum distance to 300 sampled mesh vertices & no & yes \\
    Contact-force magnitudes & 52 & $\lVert\hat{\mathbf f}\rVert\times0.01$ for each (object part, robot link) pair & no & yes \\
    Hand-presence channels & 5 & constant with two hands (all 1); kept so that the layout extends to more limbs; one enters $\mathbf z^{-i}_t$ & yes & yes \\
    \bottomrule
  \end{tabular}
\end{table*}

\paragraph{Actions.}
Following DexMachina's hybrid action, each $\mathbf a^i_t$ is clipped to $[-1,1]$ and mapped to PD targets.
The wrist targets are residuals around the reference wrist joints, and the finger targets are absolute positions within the joint limits $[\mathbf q_{\min},\mathbf q_{\max}]$,
\begin{equation}
  \begin{aligned}
    \mathbf q^{i,\text{tgt}}_{t,\text{wrist}}&=\mathbf q^i_{t,\text{wrist}}+\mathbf s_{\text{wrist}}\odot\mathbf a^i_{t,\text{wrist}},\\
    \mathbf q^{i,\text{tgt}}_{t,\text{finger}}&=\mathbf q_{\min}+\tfrac12(\mathbf q_{\max}-\mathbf q_{\min})\odot(\mathbf a^i_{t,\text{finger}}+\mathbf 1).
  \end{aligned}
  \label{eq:action}
\end{equation}
Here $\odot$ is the elementwise product, and the wrist residual scales $\mathbf s_{\text{wrist}}$ are 0.1\,m for translation and $1.0$\,rad for rotation.
Mimic joints receive their parent's target multiplied by the coupling ratio and all targets are clamped to the joint limits.
The wrist limits are the range of the reference, widened by $\pm0.2$\,m and $\pm0.5$\,rad.
The PD gains ($k_p$ / $k_v$ / force limit) of the Inspire hands are 20 / 1 / 50 for the fingers, 80 / 5 / 50 for wrist rotation and 350 / 12 / 50 for wrist translation, with a hand gravity compensation of 0.8, and \cref{tab:suppl-hands} gives those of the other hands.

\subsection{Network Architecture}
\label{sec:suppl-arch}

All layers use ELU activations.
\begin{itemize}
  \item \textbf{Actor} (shared by both hands, 111,948 parameters).
    Its mean is $\boldsymbol\mu_\theta(\mathbf o^i_t,\mathbf z^{-i}_t)=\Phi_\theta\big(E_\theta(\mathbf o^i_t),M_\theta(\mathbf z^{-i}_t)\big)$ (\cref{eq:factor}), with a local encoder $E_\theta$, a partner encoder $M_\theta$ and a trunk $\Phi_\theta$.
    The local encoder $E_\theta$ maps $157\to128$.
    The partner encoder $M_\theta$ is a Deep-Sets encoder~\cite{zaheer2017deepsets} over the partners of a hand.
    With two hands, each hand has one partner.
A per-partner encoder ($56\to64$) is applied to each partner, its outputs are summed over the partners and concatenated with their count, and an aggregator ($65\to64$) follows.
    At every step the actor is evaluated once per hand, with the roles of own and partner hand swapped.
    Finger actions mean the same for both hands, because each maps $[-1,1]$ onto the joint limits (\cref{eq:action}) and the left and right hands have the same limits.
    The separate standard deviations below let the two hands explore differently when they take different roles.
    The trunk $\Phi_\theta$ maps $192\to256\to128$, followed by a linear head $128\to12$.
    The standard deviations $\boldsymbol\sigma^i\in\mathbb R^{12}$ are learned, state-independent, separate for each hand, and initialized to $\log\boldsymbol\sigma^i=\mathbf 0$, which gives the covariance $\boldsymbol\Sigma^i=\mathrm{diag}(\boldsymbol\sigma^i)^2$ of \cref{eq:factor}.
  \item \textbf{Critic} (683,138 parameters): $499\to512\to512\to256\to128\to2$, one value head per hand.
\end{itemize}

\subsection{PPO Details}
\label{sec:suppl-ppo}

The policy loss $\mathcal L_\pi$ is the negative batch mean of the per-hand surrogate of \cref{eq:ppo}~\cite{schulman2017ppo}.
Rewards are scaled by 0.1.
For each hand, the advantage $A^i_t$ is computed with GAE~\cite{schulman2016gae} ($\gamma=0.99$, $\lambda_{\text{GAE}}=0.95$) from that hand's value head, and each hand's advantages are normalized over the whole batch.
Both value heads regress the same shared return.
The value loss is PPO's clipped value loss on normalized returns, averaged over samples and over the two heads.
The total loss is $\mathcal L=\mathcal L_\pi+\tfrac12 c_V\mathcal L_V+c_{\text{bnd}}\mathcal L_{\text{bnd}}$ with $c_V=4$ and $c_{\text{bnd}}=10^{-4}$, where the bound loss $\mathcal L_{\text{bnd}}$ penalizes action means outside $[-1.1,1.1]$.
There is no entropy bonus.
The clip range is $\epsilon=0.2$ for both the ratios and the values.
We use Adam with an initial learning rate of $3\times10^{-4}$.
The rate is adapted after each pass over the batch to keep the KL divergence near $0.008$.
It is divided by 1.5 when the KL exceeds 0.016 and multiplied by 1.5 when the KL falls below 0.004, and it stays within $[10^{-6},10^{-2}]$.
Gradients are clipped to norm 1.
Each update collects 32 steps from 12,000 parallel environments (384,000 transitions), and then makes 14 passes over the batch with minibatches of 96,000 transitions.

\subsection{Reward Terms}
\label{sec:suppl-reward}

This section defines every term of the shared reward in \cref{eq:reward}.
Its base reward is
\begin{equation}
  r_{\text{base}}=r_{\text{task}}+\lambda_{\text{imi}}r_{\text{imi}}+\lambda_{\text{bc}}r_{\text{bc}}+\lambda_{\text{con}}r_{\text{con}}-r_{\text{pen}},
  \label{eq:base-reward}
\end{equation}
where $r_{\text{task}}$ rewards tracking the object, $r_{\text{imi}}$, $r_{\text{bc}}$ and the contact matching term $r_{\text{con}}$ keep the keypoints, joint targets and contacts close to the retargeted reference ($\mathbf y^i_t$, $\mathbf q^i_t$, $\mathbf c^i_t$), and $r_{\text{pen}}$ penalizes large contact forces and actions, all as in DexMachina~\cite{mandi2026dexmachina}.
The retargeting runs a per-frame kinematic optimization and then a physical replay with the object held fixed.
All terms are computed after the physics step on the state reached by $\mathbf a_t$ and are compared with frame $t+1$ of the demonstration (clamped to $T-1$).
The exception is $r_{\text{bc}}$, which compares the PD target issued by $\mathbf a_t$ with $\mathbf q_t$.
An episode whose simulation state becomes NaN receives $r_t=-1$ and is reset.
For two orientations $\mathbf R$ and $\mathbf R'$, $d_{\text{rot}}(\mathbf R,\mathbf R')=\arccos\big((\mathrm{tr}(\mathbf R^\top\mathbf R')-1)/2\big)$ is the angle of the rotation between them.
\Cref{tab:suppl-reward} lists the weights.

\begin{table}[!tbp]
  \caption{\textbf{Reward weights.} Group A: Ketchup-300, Waffleiron-300 and Notebook-300; group B: the other four demonstrations.}
  \label{tab:suppl-reward}
  \centering
  \small
  \begin{tabular}{@{}lccccc@{}}
    \toprule
    Group & $\lambda_{\text{imi}}$ & $\lambda_{\text{con}}$ & $\lambda_{\text{bc}}$ & $\lambda_{\text{hg}}$ & $\lambda_{\text{hoc}}$ \\
    \midrule
    A & 0.3 & 3.0 & 0.3 & 0.5 & 1.0 \\
    B & 0.2 & 2.0 & 0.2 & 0.5 & 1.0 \\
    \bottomrule
  \end{tabular}
\end{table}

\paragraph{Object tracking~\cite{mandi2026dexmachina}.}
\begin{equation}
  \begin{aligned}
    r_{\text{task}}={}&\exp\!\big(-10\,\lVert\hat{\mathbf p}-\mathbf p\rVert\big)\cdot\exp\!\big(-d_{\text{rot}}(\hat{\mathbf R},\mathbf R)\big)\\
    &\cdot\exp\!\big(-2.5\,(\hat\phi-\phi)^2\big).
  \end{aligned}
\end{equation}
For an object without joints, the last factor is omitted.
For several objects, the factors are computed for each object and multiplied.

\paragraph{Motion imitation~\cite{mandi2026dexmachina}.}
Let $(\hat{\mathbf p}^i_{\mathrm{wr}},\hat{\mathbf R}^i_{\mathrm{wr}})$ and $(\mathbf p^i_{\mathrm{wr}},\mathbf R^i_{\mathrm{wr}})$ be the wrist poses of the robot and of the reference motion, and $\hat{\mathbf y}^i_k$ and $\mathbf y^i_k$ the corresponding positions of the 18 keypoints.
The imitation term is
\begin{equation}
  \begin{aligned}
    r_{\text{imi}}={}&\tfrac18\textstyle\sum_i\big(e^{-3d_{\text{rot}}(\hat{\mathbf R}^i_{\mathrm{wr}},\mathbf R^i_{\mathrm{wr}})}+e^{-10\lVert\hat{\mathbf p}^i_{\mathrm{wr}}-\mathbf p^i_{\mathrm{wr}}\rVert}\big)\\
    &+\tfrac1{72}\textstyle\sum_i\sum_{k=1}^{18}e^{-20\lVert\hat{\mathbf y}^i_k-\mathbf y^i_k\rVert}.
  \end{aligned}
\end{equation}

\paragraph{Behavior cloning~\cite{mandi2026dexmachina}.}
Let $\Delta q_j$ be the range of joint $j$.
Over all 36 joints of the two hands,
\begin{equation}
  r_{\text{bc}}=\tfrac1{36}\textstyle\sum_j\exp\!\big(-500\,(q^{\text{tgt}}_{t,j}-q_{t,j})^2/(2\,\Delta q_j)\big).
\end{equation}

\paragraph{Contact matching~\cite{mandi2026dexmachina}.}
For each hand $i$, part $m$ and robot collision link $l$ (13 per hand), $\hat{\mathbf c}^i_{ml}$ is the robot's force-weighted contact point, taken relative to the base pose of the reached object, and $\mathbf c^i_{ml}$ is the retargeted contact location, taken relative to the base pose of the demonstrated object.
The distance $d^i_{ml}$ is $\lVert\hat{\mathbf c}^i_{ml}-\mathbf c^i_{ml}\rVert$ when both the robot link and the reference link touch part $m$, and 1\,m otherwise.
The term is
\begin{equation}
  r_{\text{con}}=\tfrac14\textstyle\sum_i\sum_m\tfrac1{13}\sum_l\exp\!\big(-10\,d^i_{ml}\big).
\end{equation}

\begin{figure*}[!tp]
  \centering
  \includegraphics[width=\linewidth]{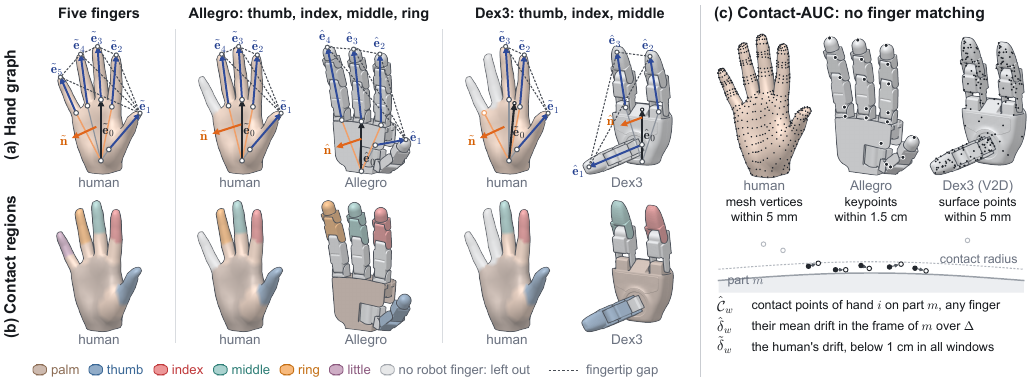}
  \caption{\textbf{Interaction rewards and Contact-AUC on hands with fewer fingers.}
    Each robot finger is matched to the human finger it is retargeted from, and the human fingers without a robot counterpart (grey) are left out of both rewards.
    (a)~The hand graph keeps the edges $\mathbf e_j$ and the fingertip gaps (dashed) of the matched fingers.
    On the human and the robot hand, the palm normal $\mathbf n$ is spanned by the index base and the outermost finger base that the robot hand has (orange lines).
    Dex3 has no finger base beyond the middle finger, so its normal is a fixed axis of the palm link, calibrated once per episode to the human normal, which keeps the little base.
    For Dex3, the human and robot palm directions $\mathbf e_0$ point to the midpoint of the index and middle bases.
    (b)~The hand-object contact reward compares the palm and the outer segments of the matched fingers.
    (c)~Contact-AUC needs no finger matching.
    Its contact points $\hat{\mathcal C}_w$ are all points of the robot hand near the part, from any finger, which are the keypoints within 1.5\,cm on the DexMachina suite and the surface points within 5\,mm on V2D.
    Their drift $\hat\delta_w$ is scored on the windows in which the human hand holds the part with a drift $\tilde\delta_w$ below 1\,cm.
    A tilde marks the human hand and a hat the robot hand.}
  \label{fig:suppl-fingers}
\end{figure*}

\paragraph{Hand graph reward.}
The definitions below are given for five-fingered hands.
For a hand with fewer fingers, each robot finger is matched to the human finger it is retargeted from (the thumb, index, middle and ring fingers for Allegro, and the thumb, index and middle fingers for Dex3).
The edges, fingertip gaps and contact regions of human fingers without a robot counterpart are left out, and \cref{fig:suppl-fingers} shows how the palm normal is adapted.
In the direction and gap errors of \cref{eq:handgraph-err}, the edges are built in the same way from the robot keypoints ($\hat{\mathbf e}$) and from the MANO~\cite{romero2017mano} joints ($\tilde{\mathbf e}$), and on a five-fingered hand the edge set is $\mathcal E=\{\mathbf n,\mathbf e_0,\dots,\mathbf e_5\}$.
For keypoint positions $\mathbf y$, the palm normal $\mathbf n$ is the unit vector along $(\mathbf y_{\text{index base}}-\mathbf y_{\text{wrist}})\times(\mathbf y_{\text{little base}}-\mathbf y_{\text{wrist}})$, the palm direction $\mathbf e_0$ is the unit vector from the wrist to the base of the middle finger, or to the midpoint of the index and middle bases on a hand without ring and little fingers, and the direction $\mathbf e_j$ of finger $j$ is the unit vector from its base to its tip.
In $\mathrm{err}_{\text{gap}}$, $\hat{\mathbf y}_j$ and $\tilde{\mathbf y}_j$ denote the tip of finger $j$.
The human edges are stored in the frame of the demonstrated object and rotated by the base orientation of the \emph{reached} object.
In $\mathrm{err}_{\text{dir}}$, the palm ($\mathbf n$, $\mathbf e_0$) and the fingers share the weight equally.
Among the fingers, the thumb has weight $0.5/4.5$ and each other finger $1/4.5$, or $0.5/(J-0.5)$ and $1/(J-0.5)$ on a hand with $J$ matched fingers.
In $\mathrm{err}_{\text{gap}}$, the pairs in $\mathcal P$, seven on a five-fingered hand, are the thumb with each other finger and each pair of neighboring fingers.
The human fingertip distances are scaled by the retargeting scale $\kappa$ of the hand, for example 1.2 for Inspire hands and 1.7 for Allegro hands.
We use $\beta_{\text{dir}}=5$ and $\beta_{\text{gap}}=20$.

\paragraph{Hand-object contact reward.}
On the human hand, a MANO vertex belongs to the region of its dominant skinning joint.
We compute $\tilde d_{g,m}$ and $\tilde{\mathbf p}_{g,m}$ from the region's closest vertex, both in the frame of the part.
On the robot, the palm region is the palm link (256 surface points), the thumb region is the two distal thumb links (64 points), and each other finger's region is its distal link (48 points).
The other hands use the same point counts on the links of \cref{tab:suppl-hands}, where the Allegro regions include the fingertip links and the Schunk palm consists of two links.
The robot's distances $\hat d_{g,m}$ and closest points $\hat{\mathbf p}_{g,m}$ are read from precomputed distance fields of the object parts, in the frame of each part, which have 3\,mm resolution and a 6\,cm margin.
When the simulator reports that one of the region's links touches the part, $\hat d_{g,m}=0$ and $\hat{\mathbf p}_{g,m}$ is the contact point of the most distal touching link.
In \cref{eq:hoc} we use $d_{\max}=5$\,cm, $d_{\text{tol}}=1$\,cm and $\beta_{\text{hoc}}=50\,\mathrm m^{-1}$, \ie a scale of 2\,cm.
Before normalization, the weight $\omega_{g,m}$ is 1 if the robot or the human region is within 1\,cm of the part, 0.5 if either is within 5\,cm, and 0 otherwise.
A hand with all weights zero scores 1.

\paragraph{Penalties~\cite{mandi2026dexmachina}.}
$r_{\text{pen}}=0.1\cdot\frac1{52}\sum_{(m,l)}\max(0,\,\lVert\hat{\mathbf f}_{ml}\rVert-500\,\mathrm N)+0.01\cdot\frac1{24}\lVert\mathbf a_t\rVert^2$, over all (object part $m$, robot link $l$) pairs, where $\hat{\mathbf f}_{ml}$ is their contact force.

\begin{table*}[!tp]
  \caption{\textbf{Settings that depend on the hand} (DexMachina suite). PD gains are $k_p$ / $k_v$ / force limit. Contact regions give the number of robot links in the palm region, the thumb region and each other finger's region. The weights are those of group A (\cref{tab:suppl-reward}); group B uses 0.2 / 2.0 / 0.2 with every hand.}
  \label{tab:suppl-hands}
  \centering
  \footnotesize
  \setlength{\tabcolsep}{8pt}
  \begin{tabular}{@{}lcccc@{}}
    \toprule
    & Inspire & Allegro & XHand & Schunk \\
    \midrule
    Fingers (fingertip gaps) & 5 (7) & 4 (5) & 5 (7) & 5 (7) \\
    Finger joints (mimic) & 12 (6) & 16 (0) & 12 (0) & 20 (11) \\
    Action & 12 & 22 & 18 & 15 \\
    Keypoints (distinct links) & 18 & 25 (21) & 18 & 33 (28) \\
    Collision links & 13 & 21 & 13 & 23 \\
    Local part $\mathbf o^i_t$ & 157 & 200 & 163 & 229 \\
    Partner part $\mathbf z^{-i}_t$ & 56 & 74 & 62 & 75 \\
    Critic input & 499 & 661 & 511 & 775 \\
    PD gains, fingers & 20 / 1 / 50 & 30 / 2 / 100 & 20 / 1.5 / 100 & 80 / 4.5 / 80 \\
    \quad wrist rotation & 80 / 5 / 50 & 60 / 5 / 100 & 60 / 3 / 100 & 80 / 5 / 100 \\
    \quad wrist translation & 350 / 12 / 50 & 350 / 20 / 100 & 350 / 15 / 100 & 350 / 8 / 100 \\
    Retargeting scale & 1.2 & 1.7 & 1.4 & 1.3 \\
    Contact regions (links) & 1 / 2 / 1 & 1 / 3 / 3 & 1 / 2 / 1 & 2 / 2 / 2 \\
    $\lambda_{\text{imi}}$ / $\lambda_{\text{con}}$ / $\lambda_{\text{bc}}$ & 0.3 / 3.0 / 0.3 & 0.2 / 2.0 / 0.2 & 0.2 / 2.0 / 0.2 & 0.2 / 2.0 / 0.3 \\
    \bottomrule
  \end{tabular}
\end{table*}

\subsection{A-RSI Details}
\label{sec:suppl-curriculum}

\begin{figure*}[!tp]
  \centering
  \includegraphics[width=\linewidth]{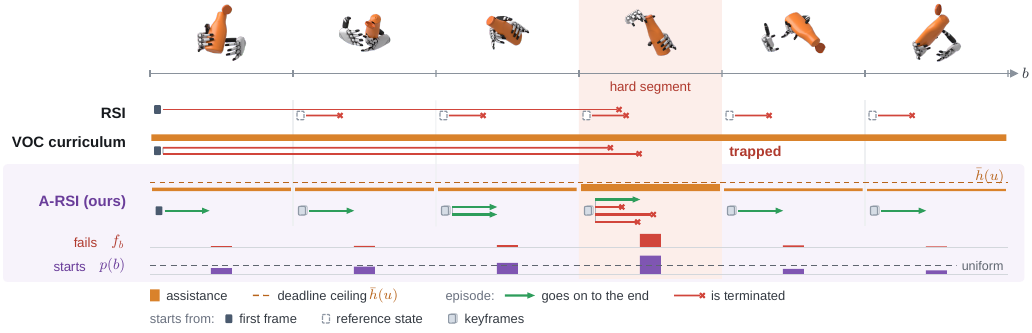}
  \caption{\textbf{How RSI, a VOC curriculum and A-RSI handle a hard segment} (illustration).
    Each line is an episode, drawn from the state it starts from (card).
    RSI~\cite{peng2018deepmimic} starts from reference states.
    The hands of these states do not hold the object, so the episodes end at once, and only the episodes that start at the first frame learn.
    A VOC curriculum~\cite{mandi2026dexmachina,zhu2026chord} holds the object with one level and lowers it only when episodes reach the end.
    Once episodes are stopped at the hard segment, the level stalls, and so does training (\cref{fig:assistance}).
    A-RSI gives each segment its own level under a deadline ceiling $\bar h(u)$.
    It starts episodes from keyframes, which are states reached in earlier episodes, and starts most of them where the policy fails, with probability $p(b)$ computed from the failure rates $f_b$ (\cref{eq:startdist}).}
  \label{fig:arsi}
\end{figure*}

\paragraph{Virtual object controller.}
As in DexMachina~\cite{mandi2026dexmachina}, for each of the 7 object DoFs (base position, base rotation and joint angle), the VOC applies
\begin{equation}
  \mathbf f^{\text{voc}}=\clip\big(k^{\text{voc}}_p\,\Delta\mathbf x_t-k^{\text{voc}}_v\,\dot{\hat{\mathbf x}}_t,\ -f_{\max},\ f_{\max}\big),
\end{equation}
where $\Delta\mathbf x_t$ is the error to the next demonstrated state, the difference in position and the joint angle and the rotation vector of $\mathbf R_{t+1}\hat{\mathbf R}_t^\top$ for the orientation.
At reset, an episode that starts in segment $b$ draws the gains of each DoF as $k^{\text{voc}}_p=\eta\,h_b$ with $\eta\sim\mathcal U[\eta_{\min},1]$, and sets $k^{\text{voc}}_v=k^{\text{voc}}_p/16$ and $f_{\max}=50\cdot\ind[h_b>0]$.

\paragraph{Segments.}
Segment $b$ covers the frames $\mathcal T_b=[15b,\ \min(15(b+1),T-10))$ for $b=0,\dots,B-1$, with $B=\lceil(T-10)/15\rceil$.
Reference-state starts lie in frames $0,\dots,T-11$ and keyframe starts at most at frame $T-10$, so every episode has at least 10 steps.
This gives 20 segments for the 300-frame demonstrations, 13 for Box-200, 11 for Mixer-170 and 6 for Ketchup-100.

\paragraph{Success and failure.}
An episode succeeds if it reaches the end of the demonstration, and fails if it is terminated early or its state becomes NaN.
Losing the object does not by itself end an episode, so success means that the tracking threshold of \cref{eq:termination} was met until the end of the demonstration.
When $N$ episodes that started in segment $b$ end at the same step, the moving-average failure rate $f_b$ moves toward their failure share with weight $1-(1-0.02)^N$.
This weight gives the batch the same total weight as $N$ per-episode moving-average steps of size 0.02.

\paragraph{Assistance updates.}
A segment is updated at PPO update $u$ when $u\ge u_0$, it has at least 64 finished episodes since its last update, and at least 40 updates have passed since its assistance last changed.
Its episode counts are reset each time it is updated.
A level that decays below the smallest nonzero level $h_{\min}$ is set to zero.
Restoring assistance from zero sets $h_b=h_{\min}$, and assistance never exceeds $h_{\max}$.
The deadline ceiling is
\begin{equation}
  \bar h(u)=\begin{cases}h_{\max}\,(h_{\min}/h_{\max})^{u/D} & u<D,\\ h_{\min}\max\big(0,\,1-(u-D)/(U-D)\big) & u\ge D.\end{cases}
\end{equation}
When $\bar h(u)>0$, any segment with $h_b>\bar h(u)/\alpha$ is set to $\bar h(u)$, so a level stays at most one decay step above the ceiling.
When $\bar h(u)=0$, every segment is set to zero.

\paragraph{Keyframes.}
Each segment $b>0$ keeps a ring buffer of up to 256 keyframes.
A keyframe holds the hands' and the object's joint positions and velocities, the hands' PD targets, the last action, and the assistance under which it was reached.
At a given step, an episode is a capture candidate if it started at frame 0 or from a keyframe, is at least 15 steps in, and is at least 10 frames before the end of the demonstration in a segment $b>0$.
Its tracking reward must also be at least 0.3 at this step, and its mean tracking reward over the episode at least 0.5.
A candidate state is picked with probability 0.002 per step.
It is stored only if its episode then reaches the end of the demonstration with a mean tracking reward of at least 0.5 from the pick onward.
Storing states only from episodes that go on to the end prevents the moments just before a failure from becoming restart points that reproduce it.
A keyframe is usable while its recorded assistance is at most $h_b$.
When segment $b>0$ holds at least 32 usable keyframes, a start in $b$ restores one of them, chosen uniformly.
The episode's assistance is still the segment's current $h_b$.

\paragraph{Termination, stopping and checkpoint.}
Following DexMachina, an episode that starts at frame $t_0$ is terminated at step $n$ once its accumulated tracking reward falls below $r_{\min}$ per step,
\begin{equation}
  \textstyle\sum_{j<n}r_{\text{task},j}+G\,\ind[t_0>0]<r_{\min}\,n ,
  \label{eq:termination}
\end{equation}
where the grace credit $G$ covers the first steps of mid-sequence starts.
These episodes begin in a state that they did not produce, and without the credit, many of them would be terminated before they could recover.
We use $r_{\min}=0.5$ and $G=8$, and the test is applied every 5 steps.
There is no early termination at evaluation.
Training stops at the update cap of the demonstration, or once every segment has been unassisted for a cumulative 1,500 updates, whichever comes first.
Both conditions are checked every 10 updates.
The checkpoint score is the tracking return $\sum_t r_{\text{task},t}$ of an episode, averaged over the last 100 finished training episodes.
A new best must beat the previous one by more than 1.0 and can only be set from update 10 on.
The best-score record and the 100-episode window are reset whenever the assistance of any segment changes.

\Cref{alg:curriculum} summarizes the procedure, \cref{tab:suppl-curriculum} lists its hyperparameters, and \cref{tab:suppl-clips} lists the settings of each demonstration.

\begin{algorithm}[!htbp]
  \caption{Assisted reference-state initialization (A-RSI), one demonstration}
  \label{alg:curriculum}
  \small
  \begin{algorithmic}[1]
    \State $h_b\gets h_{\max}$, $f_b\gets0$, keyframes $\mathcal K_b\gets\emptyset$ for all $b$; $Z\gets0$ (updates with all $h_b=0$)
    \For{update $u=1,2,\dots$}
      \If{$u\ge u_0$}
        \For{each ready segment $b$ with success rate $s_b$}
          \State update $h_b$ by \cref{eq:help}
        \EndFor
      \EndIf
      \State lower every segment with $h_b>\bar h(u)/\alpha$ to $\bar h(u)$
      \State \textbf{if} all $h_b=0$ \textbf{then} $Z\gets Z+1$
      \State \textbf{if} any $h_b$ changed \textbf{then} reset the checkpoint record
      \For{each environment that resets}
        \State score the finished episode against its start segment; update $f_b$
        \State draw $b\sim p$ (\cref{eq:startdist}); with probability $\lambda_0$, $b\gets0$
        \If{$b=0$}
          \State start at frame 0
        \ElsIf{$\mathcal K_b$ holds $\ge32$ usable keyframes}
          \State restore a uniformly chosen usable keyframe
        \Else
          \State reference-state start at $t_0\sim\mathcal U(\mathcal T_b)$
        \EndIf
        \State draw the VOC gains from $h_b$
      \EndFor
      \State collect rollouts; terminate by \cref{eq:termination}; capture and store keyframes
      \State per-hand PPO update (\cref{eq:ppo})
      \State save the checkpoint if its training score is a new best
      \State \textbf{if} $u$ reaches the cap \textbf{or} $Z>1500$ \textbf{then} stop
    \EndFor
  \end{algorithmic}
\end{algorithm}

\begin{table}[!htbp]
  \caption{\textbf{A-RSI hyperparameters}, shared by all demonstrations unless listed in \cref{tab:suppl-clips}.}
  \label{tab:suppl-curriculum}
  \centering
  \small
  \begin{tabular}{@{}p{0.62\linewidth}l@{}}
    \toprule
    Setting & Value \\
    \midrule
    Max.\ assistance $h_{\max}$; $k^{\text{voc}}_v/k^{\text{voc}}_p$; force limit $f_{\max}$ & 80; 1/16; 50 \\
    Lower gain bound $\eta_{\min}$ & \cref{tab:suppl-clips} \\
    Assistance step $\alpha$; snap to zero $h_{\min}$ & 0.9; 0.05 \\
    Success thresholds $s_{\text{hi}}$ / $s_{\text{lo}}$ & 0.8 / 0.4 \\
    Episodes per update; updates between changes & 64; 40 \\
    Wait $u_0$; deadline $D$; zero point $U$ & \cref{tab:suppl-clips} \\
    Segment length; end margin & 15; 10 frames \\
    Uniform share $\lambda_{\text{u}}$; backward credit $\gamma_{\text{seg}}$; failure-rate step & 0.3; 0.75; 0.02 \\
    Probability of a forced start at frame 0, $\lambda_0$ & 0.25 \\
    Keyframes: capacity; capture prob.; min.\ usable & 256; 0.002; 32 \\
    Grace credit $G$; threshold $r_{\min}$; check grid & 8; 0.5; 5 steps \\
    Unassisted updates before stopping & 1,500 (cumulative) \\
    \bottomrule
  \end{tabular}
\end{table}

\begin{table*}[!tp]
  \caption{\textbf{Settings of each demonstration} on the DexMachina suite. $T$ is the length in frames and $B$ the number of segments. Assistance is zero in every segment by update $U$. Two groups of settings are used, one for Ketchup-300, Waffleiron-300 and Notebook-300 and one for the other four demonstrations. On the SPIDER and ManipTrans suites, one setting is used for all demonstrations (\cref{sec:suppl-impl}). The groups, with their reward weights, lower gain bounds $\eta_{\min}$ and waits $u_0$, and the update caps are those of DexMachina's released Inspire runs on the same demonstrations.}
  \label{tab:suppl-clips}
  \centering
  \small
  \begin{tabular}{@{}lcccccccc@{}}
    \toprule
    Demonstration (ARCTIC sequence, frames) & $T$ & $B$ & Update cap & $\eta_{\min}$ & $u_0$ & $D$ & $U$ & $\lambda_{\text{imi}}$ / $\lambda_{\text{con}}$ / $\lambda_{\text{bc}}$ \\
    \midrule
    Ketchup-300 (s01 u02, 40--340)    & 300 & 20 & 13,000 & 0.5   & 100 & 9,000 & 10,000 & 0.3 / 3.0 / 0.3 \\
    Waffleiron-300 (s01 u01, 40--340) & 300 & 20 & 13,000 & 0.5   & 100 & 9,000 & 10,000 & 0.3 / 3.0 / 0.3 \\
    Notebook-300 (s02 u02, 40--340)   & 300 & 20 & 13,000 & 0.5   & 100 & 9,000 & 10,000 & 0.3 / 3.0 / 0.3 \\
    Mixer-300 (s01 u01, 40--340)      & 300 & 20 & 10,000 & 0.6   & 200 & 6,920 & 7,690  & 0.2 / 2.0 / 0.2 \\
    Ketchup-100 (s01 u01, 30--130)    & 100 & 6  & 5,000  & 0.6   & 200 & 3,460 & 3,840  & 0.2 / 2.0 / 0.2 \\
    Mixer-170 (s01 u01, 30--200)      & 170 & 11 & 5,000  & 0.6   & 200 & 3,460 & 3,840  & 0.2 / 2.0 / 0.2 \\
    Box-200 (s01 u01, 30--230)        & 200 & 13 & 5,000  & 0.6   & 200 & 3,460 & 3,840  & 0.2 / 2.0 / 0.2 \\
    \bottomrule
  \end{tabular}
\end{table*}

\begin{figure*}[!tp]
  \centering
  \includegraphics[width=\linewidth]{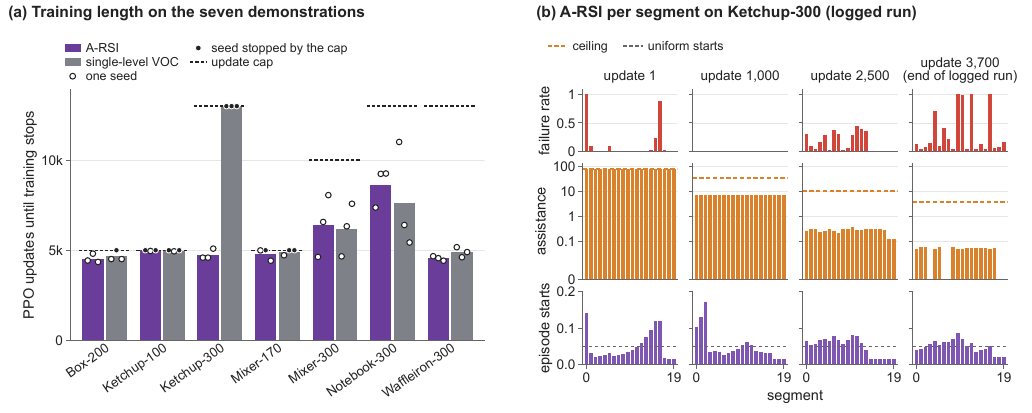}
  \caption{\textbf{Training length and per-segment assistance.} (a)~PPO updates until training stops on the DexMachina suite, for DexDuet with A-RSI and with DexMachina's single-level VOC curriculum in its place (rows (a) and (g) of \cref{tab:ablation}, with the same policy, rewards and stopping rule). A-RSI trains much shorter on Ketchup-300, where the single level stalls. (b)~Failure rate $f_b$ of each segment $b$ (top), its assistance $h_b$ under the ceiling $\bar h(u)$ (middle) and the probability $p(b)$ that an episode starts in it (bottom), in a separate Ketchup-300 run with the configuration of row (a). A-RSI first lowers the assistance of all segments together and then segment by segment, lowering it where segments succeed and raising it again where they fail, and episodes start more often in and just before the segments that fail.}
  \label{fig:assistance}
\end{figure*}

\paragraph{Comparison with DexMachina's schedule.}
Both use the same VOC, gain magnitudes and early-termination rule.
DexMachina keeps one assistance level for all episodes.
The level falls when every episode-reward average exceeds its threshold and the mean episode length is close to $T$.
At every decrease, DexMachina redraws the gains of all environments in the middle of their episodes.
Every episode starts at frame 0.
A-RSI keeps one level per segment, and this level falls or rises with the segment's success rate under the deadline ceiling.
Its gains are drawn once per episode, and its starts are spread over the demonstration.
DexMachina reaches zero assistance through a hard stop near the end of the update budget, whereas A-RSI gradually reduces assistance throughout training.
On the DexMachina suite, A-RSI reaches zero assistance by update $U\approx0.77\times$ the cap.

\section{Benchmarks and Evaluation}
\label{sec:suppl-bench}

The four benchmarks differ in their demonstrations, hands and simulators, and on each benchmark all methods run in the same simulator.

\subsection{Benchmarks}

\paragraph{DexMachina suite.}
The suite consists of seven ARCTIC demonstrations of five articulated objects, namely Box-200, Ketchup-100, Ketchup-300, Mixer-170, Mixer-300, Notebook-300 and Waffleiron-300, where the number is the length in frames.
Each demonstration is performed by two Inspire, Allegro, XHand or Schunk hands.
The simulator is Genesis~\cite{genesis2024}, rigid-body, with a time step of $1/60$\,s (one control step per demonstration frame), 2 substeps, the Newton solver, gravity and self-collision groups.

\paragraph{SPIDER suite.}
The suite consists of the seven ARCTIC demonstrations on which SPIDER is compared with DexMachina, namely Box, Ketchup-1, Ketchup-2, Laptop, Mixer, Notebook and Waffleiron (subject s01, use\_01 or use\_02, frames 30--230).
It uses two Inspire hands in Genesis.

\begin{table}[!htbp]
  \caption{\textbf{Demonstrations of the ManipTrans suite} (OakInk2 validation set, sequence@primitive). Each is the 200-frame window on which both methods are trained and evaluated.}
  \label{tab:suppl-maniptrans-clips}
  \centering
  \footnotesize
  \begin{tabular}{@{}llc@{}}
    \toprule
    Name & Sequence & Frames \\
    \midrule
    Gate-a  & d78fa@0  & 7--207 \\
    Gate-b  & e76b2@8  & 46--246 \\
    Gate-c  & e76b2@12 & 165--365 \\
    Gate-d  & e76b2@14 & 15--215 \\
    Gate-e  & 99f9d@0  & 2--202 \\
    Gate-f  & 99f9d@2  & 49--249 \\
    Gate-g  & 99f9d@7  & 0--200 \\
    Laptop  & 25281@0  & 101--301 \\
    Paste-a & f9335@0  & 0--200 \\
    Paste-b & f9335@4  & 0--200 \\
    \bottomrule
  \end{tabular}
\end{table}

\paragraph{ManipTrans suite.}
The suite consists of the ten OakInk2 demonstrations of \cref{tab:suppl-maniptrans-clips}, in ManipTrans's simulator Isaac Gym~\cite{makoviychuk2021isaacgym} and its bimanual scene with a time step of $1/60$\,s and 2 substeps.
This scene models each articulated object as two free rigid bodies.
We connect the two by a revolute joint at the calibrated pivot and axis, without damping or friction and with limits $5^\circ$ beyond the range of the demonstration, and train both ManipTrans and DexDuet in this scene with 4,096 parallel environments.
The hands are ManipTrans's Inspire models with free root bodies, gravity off and a linear and angular damping of 20.
Their finger joints are position-driven with a stiffness of 500 and a damping of 30.
PhysX uses the TGS solver with 8 position and 1 velocity iterations, a contact offset of 5\,mm and a rest offset of 0.
The friction coefficient is 4.0 on the hands, 2.0 on the object parts and 0.1 on the table.
The mass of each part is capped at 0.5\,kg.
ManipTrans moves each wrist by applying a force of at most 8.33\,N and a torque of at most 0.333\,N\,m per axis to the hand's root.
DexDuet keeps its own PD law on its six virtual wrist joints, applies the result as a force and a torque on the root, and caps them at the same values.
Isaac Gym's GPU pipeline reports only the net contact force on each body.
DexDuet therefore attributes the force on a hand link to the object part whose surface lies within 1.5\,cm of the link, and uses this attribution in its contact observations and rewards in place of Genesis's per-pair contacts.
ManipTrans keeps the schedule on gravity and object friction that it uses during training.

\paragraph{V2D challenge.}
The challenge~\cite{nvidia2026v2d} uses 30 episodes from the challenge dataset and its official evaluator.
The episodes are simulated in MuJoCo Warp~\cite{todorov2012mujoco,mujocowarp2025} at 200\,Hz with control at 50\,Hz.
\Cref{sec:suppl-locomanip} describes how DexDuet is adapted to this humanoid.

\subsection{DexDuet on the V2D Challenge}
\label{sec:suppl-locomanip}

On the V2D challenge~\cite{nvidia2026v2d}, a Unitree G1 humanoid reproduces a person moving an object, so its Dex3 hands are carried by arms and by a body that has to stay balanced.
DexDuet keeps its three components on this robot and adapts each of them to the whole body.
The observations stay those of the challenge's recipe, because the evaluator requires their layout, and we also keep its termination and PPO settings.

\paragraph{Policy.}
The challenge's recipe acts through a pre-trained whole-body motion tracker~\cite{luo2025sonic}, which turns the reference motion into targets for the 29 joints of the legs, waist and arms.
The policy adds a correction to each of these targets and to the reference targets of the 14 finger joints.
Because the arms position the hands, each per-hand policy controls one arm and its hand, which makes 14 joints, and a third policy controls the 15 joints of the legs and waist.
The two per-hand policies share their parameters.
Unlike the floating wrists on the tabletop, the arms of the G1 are mirror-image joint chains on a common torso, and the challenge's observations are expressed in the pelvis frame.
We therefore reflect the right arm and hand across the sagittal plane of the pelvis: positions, orientations and joint angles of its inputs and outputs are mapped with the joint signs derived from the URDF, so that the shared policy sees both sides in the convention of the left.
As the two hands do in \cref{sec:method-policy}, each of the three policies observes the state and the latest joint targets of the other two.
As in \cref{sec:method-policy}, each has its own probability ratio and value head, and all three receive the same reward.
The evaluator accepts a single fully connected network, so after training we convert the three policies into one network of that format and verify that it produces the same actions.

\paragraph{Curriculum.}
A-RSI replaces the challenge's curriculum and scales the challenge's virtual object controller per segment, with segments of 25 frames.
An iteration of the challenge's PPO recipe collects 98,304 samples, against 384,000 for an update on the DexMachina suite, so we multiply A-RSI's update counts by 3.9 to give each segment the same number of samples between decisions.
Keyframes store the full simulator state of the humanoid and the object, and a keyframe requires a mean tracking reward of at least 0.6, against 0.5 on the DexMachina suite.
Episodes that start from a reference state in the middle of the demonstration keep the recipe's 50-step settling phase, at the end of which the assistance decays to the level of the segment, while keyframe and first-frame starts begin at once.
Training is capped at 20,000 iterations, the budget of the retrained CHORD.
DexDuet stops earlier on 10 of the 30 episodes, a fixed number of iterations after its curriculum has withdrawn all assistance.

\paragraph{Rewards.}
We add two interaction rewards, whose weights keep their ratio to the object-tracking reward from the DexMachina suite.
The human targets come from the SOMA~\cite{saito2026soma} hands of the demonstration and are computed once per episode, and the hand-object contact reward treats each object as one part.
Both interaction rewards use the three fingers of the Dex3 hand as in \cref{sec:suppl-reward,fig:suppl-fingers}, and the fingertip gaps of the human are scaled by 1.5, the retargeting scale of this hand.

\subsection{Baselines}
\label{sec:suppl-baselines}

\hspace*{\parindent}DexMachina is evaluated with the authors' released checkpoints.

SPIDER is rerun from its official code in our Genesis scene with five seeds per demonstration.
SPIDER optimizes each demonstration separately.
One run over the 200 frames of a demonstration takes 1.4--19.9\,min (median 7.9\,min) on one GPU, and the 35 runs take 5.8\,h in total.

ManipTrans is retrained from scratch with its official protocol in the scene with joints.
This protocol trains a residual policy per demonstration on top of the released per-hand imitators, with 4,096 environments, at most 10,000 iterations and early stopping after 1,000 iterations without improvement.
To check our ManipTrans setup, we reproduce its evaluation on the 77 OakInk2 validation sequences with the Inspire hand, following the evaluation that the authors describe in their code repository.
We obtain success rates of 57.8\% on the 38 single-hand sequences and 32.9\% on the 39 bimanual ones, against 58.1\% and 39.5\% in the paper~\cite{li2025maniptrans}.
On successful rollouts, our errors are within 0.06\,cm and $0.3^\circ$ of the reported ones.

On the V2D challenge, we retrain CHORD with the challenge's official code and both of its recipes for G1 + Dex3.
The PPO recipe keeps its settings and runs for 20,000 iterations of 4,096 environments and 24 steps, or 1.97 billion environment steps per episode.
On episode 7, the simulator occasionally returns a non-finite state that would end learning, so for this episode we reset the affected environment, as our implementation does.
The FlashSAC recipe is set for 0.35 billion environment steps, so we scale its settings that depend on the budget, namely the total number of steps, the stage boundaries of its curriculum and the checkpoint interval, to the same 1.97 billion steps and keep everything else. The challenge's leaderboard also lists an entry of CHORD, whose training budget is not published and whose policy is not released (\cref{tab:suppl-v2d}).

ConTrack~\cite{liang2026contrack} is rerun from its official code on the DexMachina suite with XHand hands.
We port its environment to the scenes of DexMachina's released XHand runs, with the object not actuated, and keep its rewards, constraint handling, PPO settings and network.
It is trained with five seeds per demonstration and DexDuet's budget, namely 12,000 environments, 32 steps per update and the same maximum number of updates per demonstration, without early stopping, and, as in ConTrack, we evaluate its final checkpoint.

\subsection{Training and Evaluation}
\label{sec:suppl-impl}

\paragraph{Seeds and stopping.}
On the DexMachina suite, DexDuet is trained with five seeds per demonstration (15, 24, 42, 66 and 113), and the ablations of \cref{tab:ablation} with three (15, 42 and 113).
On the SPIDER suite, the five seeds are 15, 24, 42, 43 and 66.
On the ManipTrans suite, DexDuet and ManipTrans are both trained with three seeds (42, 24 and 66).
On V2D, every method is trained with one seed.
On the SPIDER suite, training uses 12,000 parallel environments.
On the SPIDER and ManipTrans suites, training takes at most 15,000 updates, the assistance ceiling reaches zero at update 10,000 (a deadline of 9,000 updates plus a 1,000-update ramp), and training stops 1,500 updates after all assistance is gone.
All DexDuet runs that we train in Genesis, including the ablations, use this stopping rule.

\paragraph{Evaluation.}
Every method is evaluated from the first frame of each demonstration, without assistance and with deterministic actions.
In Genesis, each policy is evaluated on 100 parallel rollouts, with the first observation read from the simulator after the reset.
On the ManipTrans suite, every policy is evaluated on 256 rollouts per demonstration, which are the last two episodes of 128 environments.
Every rollout of both methods starts at the first frame of the 200-frame window, whereas ManipTrans's own evaluation starts up to 165 frames earlier on seven demonstrations.
ManipTrans's evaluation is stochastic, so we fix its evaluation seed, and over three evaluation seeds the standard deviation of its mean ADD-AUC and success rates is at most 0.8 points.
On the V2D challenge, the official evaluator runs one rollout per episode from the first frame.
It is not deterministic, so we report the mean of four runs.

\paragraph{Compute.}
DexDuet trains each policy for 4,350--11,500 PPO updates, which are 1.67--4.42\,B transitions on the DexMachina and SPIDER suites and 0.58--1.51\,B on the ManipTrans suite, and the ablation rows (b)--(g) stop after at most 13,000 updates (4.99\,B transitions).
On the DexMachina suite, a typical run takes about one day on one GPU (median 21.7\,h).
ManipTrans trains only a residual policy on top of its released, pre-trained per-hand imitators and stops early, so its 2.1--4.9\,h per run on the ManipTrans suite exclude the training of the imitators.

\subsection{Tracking Metrics}
\label{sec:suppl-tracking}

\paragraph{ADD-AUC.}
We use DexMachina's released code unchanged.
For each part, it samples 500 mesh vertices with a fixed seed.
For each rollout and frame, it computes the mean distance between these vertices at the reached and at the demonstrated pose.
The accuracy at a threshold is the share of (rollout, frame) pairs with an error below it, for thresholds of $1,2,\dots,9$\,cm.
The AUC is the trapezoidal area under this accuracy curve, and the reported score is the mean of the two parts' AUCs.
Following the released evaluator, the state after step $t$ is compared with frame $t-1$ of the demonstration.
On the SPIDER suite, we apply the same definition to the saved object states of every method.
We sample 500 vertices per part with seed 42, use thresholds of $1,2,\dots,9$\,cm and average the AUCs of the two parts.
The moving part is placed by the joint angle.
As in SPIDER's comparison with DexMachina, the state after each step is compared with the frame of the demonstration that this step tracks.
On SPIDER's released trajectories, this gives scores close to those SPIDER reports (\cref{sec:suppl-results}).
On the ManipTrans suite, ADD-AUC is computed over each 200-frame window.

\paragraph{Success rates and MPPE.}
In \cref{tab:main}, SP-SR, MP-SR and MPPE are computed with the evaluation code of the V2D challenge on every benchmark.
For each rollout, this code averages the position error and the geodesic rotation error of each object part over all frames, with the frame alignment of ADD-AUC.
SP-SR counts a rollout as a success if the errors averaged over the parts are at most 10\,cm and 0.5\,rad, which are the thresholds of SPIDER~\cite{pan2025spider}.
MP-SR counts it as a success if the errors of every part are below 3\,cm and $30^\circ$, which are the object thresholds of ManipTrans~\cite{li2025maniptrans}.
MPPE places six points 5\,cm from the origin of each part along its axes and averages their distance over frames, parts and rollouts.
On the ManipTrans suite, rollouts that ManipTrans's environment terminates early count as failures.
For MPPE, these rollouts keep their last pose, and those terminated before the first recorded frame are left out (3 of 7,680 rollouts).
The original criteria differ.
SPIDER's criterion uses only the root part of the object.
ManipTrans's criterion tests its object thresholds at every frame from frame 8 on.
We report these original object criteria on their own benchmarks (\cref{sec:suppl-results}).

\FloatBarrier
\section{Contact-AUC}
\label{sec:suppl-metrics}

Contact-AUC measures whether each robot hand stays on the part that the human's hand holds, without sliding on it.
DexDuet's advantage in Contact-AUC remains between runs that track the object equally well (\cref{sec:suppl-matched}), and Contact-AUC agrees with the judgments of human raters (\cref{sec:suppl-perceptual}).

\subsection{Definition}

In the contact drift of \cref{eq:drift}, $\mathbf v_{k,t}=(\mathbf X^m_t)^{-1}\mathbf y_{k,t}$ expresses the world position $\mathbf y_{k,t}$ of point $k$ in the frame of the part (in homogeneous coordinates).
For the human, $\tilde\delta_w$ uses the vertices $\tilde{\mathcal C}_w$ of the hand mesh in contact at $t$ and the demonstrated part pose.
For the robot, $\hat\delta_w$ uses its points $\hat{\mathcal C}_w$ in contact and the reached part pose $\hat{\mathbf X}^m_t$, and is infinite if no point touches.
The hold windows are
\begin{equation}
  \mathcal W=\big\{w:\ \tilde{\mathcal C}_w\neq\emptyset,\ \tilde{\mathcal C}_{w^+}\neq\emptyset,\ \tilde\delta_w<\delta_{\max}\big\},
  \label{eq:windows}
\end{equation}
where $w^+=(i,m,t+\Delta)$, so the human hand touches the part at both ends of a window, and Contact-AUC scores $\hat\delta_w$ on them.
We use $\Delta=6$ frames and $\delta_{\max}=1$\,cm, and a MANO vertex of hand $i$ is in contact with part $m$ when it is within 5\,mm of the part.
Each human contact is assigned to the nearest part.
The frame of the top part is that of the bottom part rotated by $\mathbf R_z(-\hat\phi)$.
The drift $\hat\delta_w$ is also set to infinity when the object is lost or the simulation state is invalid.
The object counts as lost when its mean ADD point error over both parts exceeds 5\,cm at $t$ or $t+\Delta$.
Each window is scored by the area under $\ind[\hat\delta_w<\tau]$ between $\tau_{\min}=0.5$\,cm and $\tau_{\max}=5$\,cm, computed on a 0.5\,cm grid with the same rule as ADD-AUC.
Contact-AUC is the mean score over all windows and rollouts.
Unlike ADD-AUC, Contact-AUC compares the state after step $t$ with frame $t$.

\subsection{Contact Points on Each Benchmark}
\label{sec:suppl-contact-points}

On the DexMachina suite, the robot contacts $\hat{\mathcal C}_w$ are the keypoints of hand $i$ within 1.5\,cm of part $m$ at $t$ that are closer to $m$ than to the other part.
The keypoints of the Inspire hand are its 18 tracked link frames, which are its 5 fingertips and 13 collision links (\cref{tab:suppl-hands} for the other hands).
For SPIDER and ManipTrans, we set the recorded hand and object states of their rollouts in our Genesis scene frame by frame and read the keypoints there.

On the V2D challenge, the windows, the human contact rule and the scoring stay the same.
Each object is one rigid part, and the human hands are SOMA meshes with 2,847 vertices per hand instead of MANO meshes.
The links of the Dex3 hand are 1.6 to 2.8\,cm thick, and their origins lie up to 2\,cm from their surfaces.
We therefore use 64 points on the surface of each of its eight collision links, and the robot contacts are the points within 5\,mm of the object, the distance at which a human contact is counted.
The evaluator saves no hand states, so we record them in four runs of it per episode.
Contact-AUC is averaged over the 29 episodes in which the human hands touch the object.
In the remaining episode, the human hands never come within 5\,mm of the object.
As on the other suites, contacts of the forearms, the torso and the legs are not counted.

\subsection{Where Contact-AUC Is Lost}
\label{sec:suppl-decomp}

Each window of $\mathcal W$ falls into one of three classes.
A window is lost if the object is lost at $t$ or $t+\Delta$.
It has no contact if the object is on track but hand $i$ has no contact point on part $m$ at $t$.
Otherwise it is touching, and its drift decides its score.
A run's Contact-AUC is therefore the share of touching windows times the mean score over them.
It is also the share of windows with the object on track times the on-track Contact-AUC, which is the mean score over these windows.
\Cref{tab:suppl-decomp} gives this split for the ablations.

\begin{table*}[!tp]
  \caption{\textbf{Where Contact-AUC is lost} in the ablations (DexMachina suite, Inspire), in \%. The three classes of windows are defined in \cref{sec:suppl-decomp}, and their shares add up to 100. Score when touching is the mean score over the touching windows, and on-track Contact-AUC scores all windows that are not lost. Mean over runs $\pm$ std over training-seed means, as in \cref{tab:ablation}. Bold marks the best value in each column.}
  \label{tab:suppl-decomp}
  \centering
  \footnotesize
  \setlength{\tabcolsep}{3.5pt}
  \begin{tabular}{@{}llccccccc@{}}
    \toprule
    & & & & \multicolumn{3}{c}{Share of windows} & Score when & On-track \\
    \cmidrule(lr){5-7}
    & Method & Runs & Contact-AUC$\uparrow$ & Lost$\downarrow$ & No contact$\downarrow$ & Touching$\uparrow$ & touching$\uparrow$ & Contact-AUC$\uparrow$ \\
    \midrule
    \multicolumn{9}{@{}l}{\textit{Ablation (DexMachina suite, Inspire, seeds 15, 42 and 113)}} \\
    (a) & DexDuet                                      & 21 & \textbf{80.5}$\pm$3.2 & \textbf{5.7}$\pm$4.2 & \textbf{3.9}$\pm$0.6 & \textbf{90.4}$\pm$3.7 & \textbf{88.8}$\pm$0.3 & \textbf{85.1}$\pm$0.2 \\
    (b) & $\lambda_{\text{hg}}=0$                      & 21 & 71.5$\pm$1.6 & 7.6$\pm$1.7 & 8.6$\pm$1.3 & 83.7$\pm$2.1 & 85.4$\pm$1.4 & 76.5$\pm$1.3 \\
    (c) & $\lambda_{\text{hoc}}=0$                     & 21 & 65.4$\pm$4.5 & 14.4$\pm$6.2 & 10.0$\pm$2.5 & 75.7$\pm$4.6 & 82.3$\pm$3.6 & 71.7$\pm$3.1 \\
    (d) & $\lambda_{\text{hg}}=\lambda_{\text{hoc}}=0$ & 21 & 49.0$\pm$5.1 & 24.9$\pm$9.1 & 14.5$\pm$2.0 & 60.6$\pm$7.4 & 76.9$\pm$3.8 & 58.5$\pm$2.9 \\
    (e) & unified policy                               & 21 & 62.1$\pm$2.6 & 21.8$\pm$3.7 & 5.5$\pm$0.8 & 72.7$\pm$3.4 & 85.2$\pm$0.8 & 78.8$\pm$1.2 \\
    (f) & (d) with single-level VOC                              & 21 & 47.6$\pm$4.5 & 21.4$\pm$7.5 & 16.8$\pm$3.3 & 61.8$\pm$4.6 & 69.1$\pm$4.7 & 52.8$\pm$3.8 \\
    (g) & single-level VOC, no A-RSI                      & 21 & 74.2$\pm$3.2 & 7.8$\pm$0.5 & 5.8$\pm$2.8 & 86.4$\pm$2.6 & 85.7$\pm$1.4 & 80.0$\pm$3.7 \\
    \bottomrule
  \end{tabular}
\end{table*}

\subsection{Contact-AUC at Matched Tracking}
\label{sec:suppl-matched}

\begin{figure*}[!tp]
  \centering
  \includegraphics[width=\linewidth]{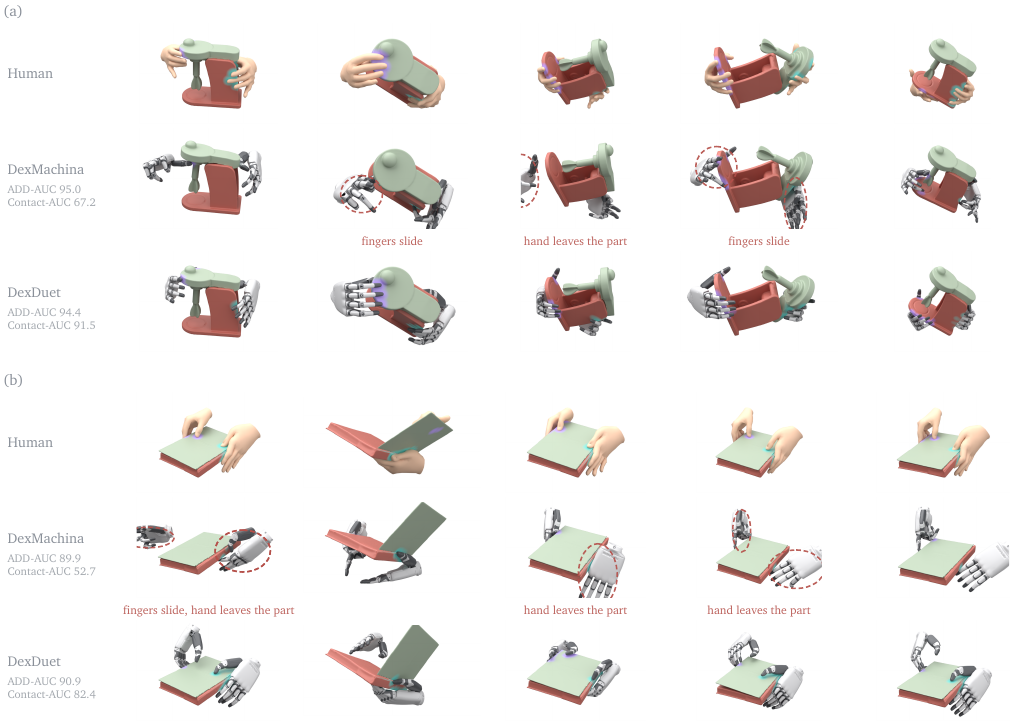}
  \caption{\textbf{Runs that track the object equally well can differ in contact preservation.} (a)~Mixer-170 and (b)~Notebook-300 with Inspire hands, each a released DexMachina~\cite{mandi2026dexmachina} checkpoint and a DexDuet run with almost the same ADD-AUC. In the marked frames, both objects are within 2\,cm of the demonstration and DexDuet preserves the human's contact, whereas DexMachina's fingers slide on the part that the human holds or its hand leaves that part (dashed circles).}
  \label{fig:suppl-matched}
\end{figure*}

At equal tracking, DexDuet preserves the human's contacts better than DexMachina and than its ablation without both interaction rewards.
To separate the contacts from tracking, we compare runs that track the object equally well.
A pair consists of two runs on the same hand and demonstration.
It is either one of DexDuet's five seeds and one of DexMachina's five released checkpoints, or DexDuet~(a) and an ablation row with the same training seed.
A pair is matched if the two ADD-AUCs differ by at most 3 points.
\Cref{tab:suppl-matched} lists the matched pairs, and \cref{fig:suppl-matched} shows two of them.

On the DexMachina suite, both methods track the object at about 94 ADD-AUC in the matched pairs.
In the median over pairs, DexDuet's Contact-AUC is higher in every hand and demonstration that has a matched pair.
The matched gap comes both from windows in which DexMachina's hand is not on the part that the human holds and from windows in which its hand touches the part but its contacts slide.

The ablation pairs keep the training seed and the demonstration fixed.
At matched tracking, removing the hand graph reward~(b) lowers Contact-AUC in 10 of 13 pairs, and removing both rewards~(d) lowers it in all 9.

\begin{table*}[!tp]
  \caption{\textbf{Contact-AUC at matched tracking} (DexMachina suite). Pairs and matching are defined in \cref{sec:suppl-matched}, and rows are named by the hand or the ablation row. ADD-AUC and Contact-AUC are differences, the first method minus the second, in points, as medians over pairs, and Contact-AUC has a 95\% bootstrap interval over demonstrations. Pairs: number of pairs, with the number of hand and demonstration cells. Higher: pairs in which the first / second method has the higher Contact-AUC. Bold marks the Contact-AUC differences whose interval excludes zero.}
  \label{tab:suppl-matched}
  \centering
  \footnotesize
  \setlength{\tabcolsep}{3.5pt}
  \begin{tabular}{@{}lcccc@{}}
    \toprule
    Comparison & Pairs & ADD-AUC & Contact-AUC [95\% CI] & Higher \\
    \midrule
    Inspire             & 28 (5)   & $+0.9$ & $\mathbf{+22.4}$ [17.6, 26.1] & 28 / 0 \\
    Allegro             & 42 (5)   & $+0.9$ & $\mathbf{+20.0}$ [10.4, 27.1] & 41 / 1 \\
    XHand               & 20 (3)   & $+1.4$ & $\mathbf{+12.4}$ [1.2, 23.0] & 19 / 1 \\
    Schunk              & 32 (3)   & $+0.9$ & $\mathbf{+16.8}$ [9.7, 32.6] & 29 / 3 \\
    All four hands      & 122 (16) & $+1.0$ & $\mathbf{+17.9}$ [13.4, 23.0] & 117 / 5 \\
    (a) vs.\ (b)        & 13 (6)   & $+1.0$ & $+7.0$ [$-1.0$, 10.5] & 10 / 3 \\
    (a) vs.\ (c)        & 8 (6)    & $-1.1$ & $+10.7$ [$-0.6$, 16.0] & 6 / 2 \\
    (a) vs.\ (d)        & 9 (5)    & $-2.1$ & $\mathbf{+18.9}$ [16.3, 24.8] & 9 / 0 \\
    (a) vs.\ (f)        & 8 (5)    & $-1.0$ & $\mathbf{+20.4}$ [16.4, 38.9] & 8 / 0 \\
    \bottomrule
  \end{tabular}
\end{table*}

\subsection{Perceptual Study}
\label{sec:suppl-perceptual}

We test whether people see the difference that Contact-AUC measures, on pairs of runs that track the object equally well.

Each rater watches 26 videos on a web page.
A video shows the human demonstration above two robot rollouts, labelled A and B, over the full demonstration at real-time speed, and the rater answers ``Which robot handles the object the way the human does?''
All robots are drawn with the same materials and without method names.
A and B are assigned at random for each video, and each rater sees the videos in one fixed random order or its reverse.
Taking part is voluntary and anonymous.
The first page tells each rater that no name or email is collected, that they may stop at any time and that the answers are used in aggregate in a research paper, and the videos start only after the rater confirms being at least 18 years old and agreeing to take part.
None of the raters is involved in this work, and no personal data are collected.

Twenty-four videos show pairs of runs on the same hand and demonstration whose ADD-AUCs differ by at most 3 points and whose Contact-AUCs differ by at least 10 points.
They are drawn with a fixed seed from the pairs of \cref{sec:suppl-matched}.
Sixteen are pairs of DexDuet and DexMachina, four per hand, and eight are pairs of DexDuet~(a) and ablation row~(b), (c) or~(d) with Inspire hands.
In every pair, the run with the higher Contact-AUC is DexDuet or row~(a).
In the eight ablation pairs, both runs are DexDuet policies with the same architecture and curriculum that differ only in the interaction rewards.
The two remaining videos are attention checks in which one robot loses the object, and raters who choose that robot are excluded.

None of the eleven raters failed an attention check.
The raters chose the run with the higher Contact-AUC in 251 of 264 judgments (95.1\%, with a 95\% interval of 88.6\% to 99.2\% from resampling raters and videos), and in each of the 24 videos most raters chose it (\cref{tab:suppl-perceptual}).
In a logistic regression of the choice on both differences, each standardized, Contact-AUC has a large positive coefficient (3.79, standard error 0.45) and ADD-AUC a small negative one ($-0.61$, standard error 0.27).
Contact-AUC therefore agrees with how people judge whether a robot handles the object the way the human does.

\begin{table}[!htbp]
  \caption{\textbf{Perceptual study.} Each video is a pair of runs that track the object about equally well. $|\Delta$ADD$|$ and $\Delta$Contact: median differences between the two runs of a pair, in points. Chosen: share of judgments that pick the run with the higher Contact-AUC, in \%. Majority: videos in which more than half of the raters pick this run. Rows (b), (c) and (d) are those of \cref{tab:ablation}, with Inspire hands. Bold marks the overall result.}
  \label{tab:suppl-perceptual}
  \centering
  \footnotesize
  \setlength{\tabcolsep}{3pt}
  \begin{tabular}{@{}lccccc@{}}
    \toprule
    Pairs & Videos & $|\Delta$ADD$|$ & $\Delta$Contact & Chosen & Majority \\
    \midrule
    \multicolumn{6}{@{}l}{\textit{DexDuet vs DexMachina}} \\
    Inspire & 4 & 1.45 & 18.1 & 100.0 & 4 / 4 \\
    Allegro & 4 & 0.90 & 15.5 & 97.7 & 4 / 4 \\
    XHand   & 4 & 1.99 & 16.1 & 95.5 & 4 / 4 \\
    Schunk  & 4 & 1.07 & 17.2 & 95.5 & 4 / 4 \\
    All     & 16 & 1.39 & 17.8 & 97.2 & 16 / 16 \\
    \midrule
    \multicolumn{6}{@{}l}{\textit{DexDuet (a) vs ablation rows}} \\
    (b) & 3 & 2.68 & 16.2 & 87.9 & 3 / 3 \\
    (c) & 3 & 1.20 & 16.0 & 87.9 & 3 / 3 \\
    (d) & 2 & 2.31 & 18.9 & 100.0 & 2 / 2 \\
    All     & 8 & 1.83 & 16.1 & 90.9 & 8 / 8 \\
    \midrule
    All pairs & 24 & 1.46 & 17.3 & \textbf{95.1} & \textbf{24 / 24} \\
    \bottomrule
  \end{tabular}
\end{table}

\FloatBarrier
\section{Additional Experimental Results}
\label{sec:suppl-results}

This section gives the per-demonstration results behind \cref{tab:main}, the test of holding after the last frame, the ablations of \cref{tab:ablation} and the real-world replays of \cref{sec:real}.

\subsection{Qualitative Results}

\begin{figure*}[!tp]
  \centering
  \includegraphics[width=\linewidth]{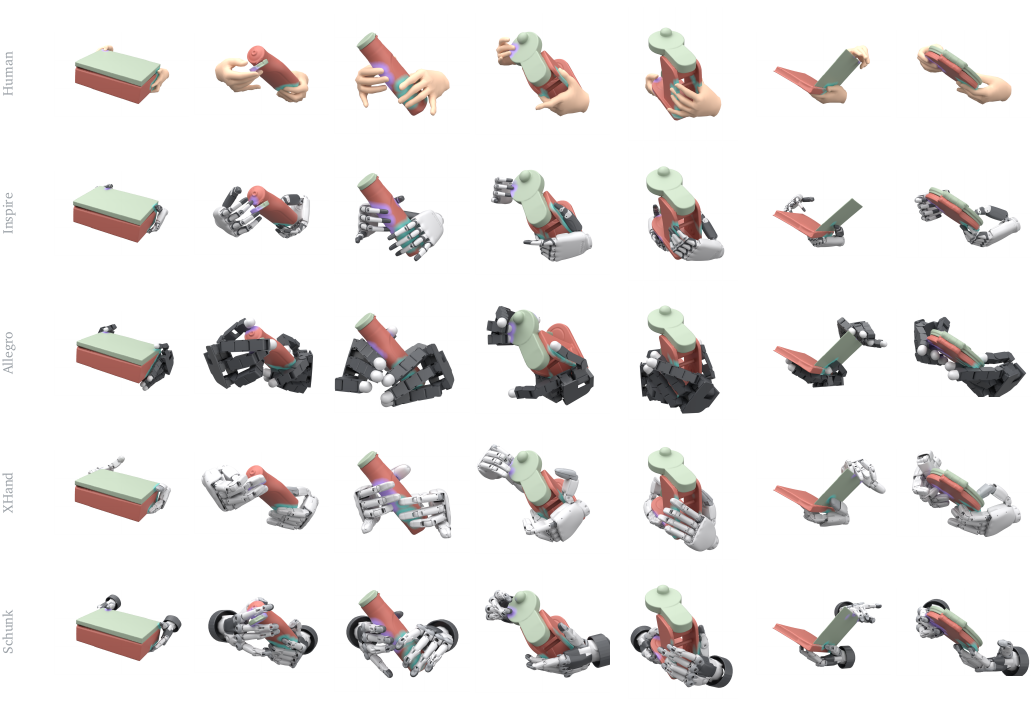}
  \caption{\textbf{DexDuet on the DexMachina suite.} Columns, from left, are Box-200, Ketchup-100, Ketchup-300, Mixer-170, Mixer-300, Notebook-300 and Waffleiron-300, and rows show the human demonstration and DexDuet with Inspire, Allegro, XHand and Schunk hands. Each robot row is the seed with the median Contact-AUC, each column is shown at the human hold with both hands closest to the middle of the demonstration from one camera, and soft patches mark where the left (teal) and right (violet) hand touch the object.}
  \label{fig:suppl-gallery-dexmachina}
\end{figure*}

\begin{figure*}[!tp]
  \centering
  \includegraphics[width=\linewidth]{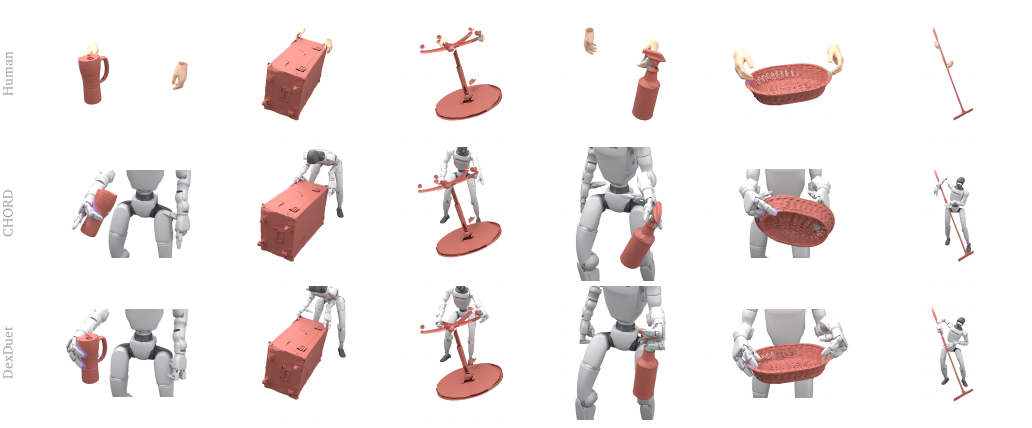}
  \caption{\textbf{CHORD and DexDuet on V2D} with the Unitree G1 and its Dex3 hands. Columns, from left, are episodes 3 (mug), 5 (box), 14 (round table), 16 (spray bottle), 27 (basket) and 28 (squeegee), and rows show the human with SOMA hands, CHORD~\cite{zhu2026chord} trained with PPO and DexDuet, each in its first scored rollout at the same training budget. Each column is shown at the human hold closest to the middle of the episode from one camera, and at that moment CHORD has lost the object in every column except the box.}
  \label{fig:suppl-gallery-v2d}
\end{figure*}

\Cref{fig:qualitative} compares the baselines and DexDuet on four pairs of runs of the DexMachina suite that track the object equally well and on two V2D episodes, and \cref{fig:suppl-matched} follows two such pairs over a demonstration.
\Cref{fig:suppl-gallery-dexmachina} shows DexDuet with all four hands on the seven demonstrations of the DexMachina suite, and \cref{fig:suppl-gallery-v2d} compares CHORD and DexDuet on six V2D episodes.

\subsection{Results per Benchmark}

\paragraph{DexMachina suite.}
\Cref{tab:suppl-dexmachina} gives the per-demonstration results with Inspire hands.
DexDuet has the higher Contact-AUC on all seven demonstrations and the higher mean ADD-AUC over them (88.7 against 52.5).
\Cref{sec:suppl-matched} compares the runs of the two methods that track the object equally well.
\Cref{tab:suppl-dexmachina-allegro,tab:suppl-dexmachina-xhand,tab:suppl-dexmachina-schunk} give the same results for the other three hands.
With each of them, DexDuet has the higher Contact-AUC on all seven demonstrations and the higher mean ADD-AUC.
With XHand hands, ConTrack~\cite{liang2026contrack} reaches a mean ADD-AUC of 30.7 and a mean Contact-AUC of 14.4, against 60.7 and 38.2 for DexMachina and 93.6 and 79.5 for DexDuet.

\begin{table*}[!tp]
  \caption{\textbf{Per-demonstration results on the DexMachina suite} (ARCTIC, two Inspire hands, Genesis), in \%, and MPPE in cm. ADD-AUC and Contact-AUC: mean $\pm$ std over five training seeds (DexDuet) or the five released checkpoints (DexMachina). SP-SR, MP-SR and MPPE (evaluation code of the challenge): means. Bold marks the best value of each metric in each column.}
  \label{tab:suppl-dexmachina}
  \centering
  \footnotesize
  \setlength{\tabcolsep}{3pt}
  \begin{tabular}{@{}llcccccccc@{}}
    \toprule
    & Method & Box-200 & Ketchup-100 & Ketchup-300 & Mixer-170 & Mixer-300 & Notebook-300 & Waffleiron-300 & Mean \\
    \midrule
    \multirow{2}{*}{ADD-AUC}
      & DexMachina          & 88.4$\pm$4.6 & 70.3$\pm$25.4 & 4.9$\pm$1.8 & \textbf{93.1}$\pm$2.3 & 62.6$\pm$31.6 & 38.3$\pm$46.7 & 10.0$\pm$7.0 & 52.5 \\
      & DexDuet             & \textbf{95.6}$\pm$1.4 & \textbf{84.1}$\pm$12.6 & \textbf{86.9}$\pm$2.1 & 86.8$\pm$18.3 & \textbf{92.1}$\pm$3.9 & \textbf{84.5}$\pm$4.8 & \textbf{90.9}$\pm$1.4 & \textbf{88.7} \\
    \midrule
    \multirow{2}{*}{Contact-AUC}
      & DexMachina          & 58.3$\pm$14.4 & 39.4$\pm$19.4 & 0.5$\pm$0.3 & 60.3$\pm$11.1 & 48.0$\pm$26.7 & 23.3$\pm$30.1 & 5.4$\pm$2.1 & 33.6 \\
      & DexDuet             & \textbf{92.1}$\pm$1.8 & \textbf{70.8}$\pm$20.0 & \textbf{78.4}$\pm$4.6 & \textbf{79.3}$\pm$16.2 & \textbf{86.1}$\pm$7.7 & \textbf{77.3}$\pm$7.0 & \textbf{73.0}$\pm$1.1 & \textbf{79.6} \\
    \midrule
    \multirow{2}{*}{SP-SR}
      & DexMachina          & 99.0 & 79.2 & 0.0 & \textbf{99.0} & 59.4 & 39.6 & 0.0 & 53.7 \\
      & DexDuet             & \textbf{100.0} & \textbf{80.0} & \textbf{100.0} & 80.0 & \textbf{100.0} & \textbf{100.0} & \textbf{100.0} & \textbf{94.3} \\
    \midrule
    \multirow{2}{*}{MP-SR}
      & DexMachina          & 99.0 & 79.2 & 0.0 & \textbf{99.0} & 59.4 & 39.6 & 0.0 & 53.7 \\
      & DexDuet             & \textbf{100.0} & \textbf{80.0} & \textbf{100.0} & 80.0 & \textbf{100.0} & \textbf{100.0} & \textbf{100.0} & \textbf{94.3} \\
    \midrule
    \multirow{2}{*}{MPPE}
      & DexMachina          & 1.06 & 14.81 & 112.96 & \textbf{1.06} & 25.33 & 50.81 & 78.59 & 40.66 \\
      & DexDuet             & \textbf{0.81} & \textbf{2.39} & \textbf{1.65} & 1.54 & \textbf{1.06} & \textbf{1.74} & \textbf{1.33} & \textbf{1.50} \\
    \bottomrule
  \end{tabular}
\end{table*}

\begin{table*}[!tp]
  \caption{\textbf{Per-demonstration results on the DexMachina suite with Allegro hands}, as in \cref{tab:suppl-dexmachina}.}
  \label{tab:suppl-dexmachina-allegro}
  \centering
  \footnotesize
  \setlength{\tabcolsep}{3pt}
  \begin{tabular}{@{}llcccccccc@{}}
    \toprule
    & Method & Box-200 & Ketchup-100 & Ketchup-300 & Mixer-170 & Mixer-300 & Notebook-300 & Waffleiron-300 & Mean \\
    \midrule
    \multirow{2}{*}{ADD-AUC}
      & DexMachina          & 91.0$\pm$2.9 & 81.2$\pm$29.3 & 71.9$\pm$33.5 & 80.0$\pm$32.3 & 72.0$\pm$31.9 & 24.9$\pm$35.7 & 39.7$\pm$43.9 & 65.8 \\
      & DexDuet             & \textbf{97.5}$\pm$0.5 & \textbf{93.1}$\pm$2.5 & \textbf{92.6}$\pm$1.7 & \textbf{96.7}$\pm$1.9 & \textbf{98.2}$\pm$0.8 & \textbf{89.7}$\pm$1.0 & \textbf{95.7}$\pm$0.3 & \textbf{94.8} \\
    \midrule
    \multirow{2}{*}{Contact-AUC}
      & DexMachina          & 38.1$\pm$1.5 & 49.3$\pm$25.3 & 51.6$\pm$25.3 & 47.9$\pm$27.8 & 45.0$\pm$23.9 & 7.0$\pm$10.1 & 12.0$\pm$13.8 & 35.8 \\
      & DexDuet             & \textbf{43.9}$\pm$4.5 & \textbf{82.4}$\pm$4.9 & \textbf{79.9}$\pm$2.5 & \textbf{72.6}$\pm$5.6 & \textbf{71.0}$\pm$3.1 & \textbf{55.8}$\pm$6.0 & \textbf{70.1}$\pm$3.9 & \textbf{68.0} \\
    \midrule
    \multirow{2}{*}{SP-SR}
      & DexMachina          & 99.0 & 79.2 & 79.2 & 79.2 & 79.2 & 19.8 & 39.6 & 67.9 \\
      & DexDuet             & \textbf{100.0} & \textbf{100.0} & \textbf{100.0} & \textbf{100.0} & \textbf{100.0} & \textbf{100.0} & \textbf{100.0} & \textbf{100.0} \\
    \midrule
    \multirow{2}{*}{MP-SR}
      & DexMachina          & 99.0 & 79.2 & 79.2 & 79.2 & 79.2 & 19.8 & 39.6 & 67.9 \\
      & DexDuet             & \textbf{100.0} & \textbf{100.0} & \textbf{100.0} & \textbf{100.0} & \textbf{100.0} & \textbf{100.0} & \textbf{100.0} & \textbf{100.0} \\
    \midrule
    \multirow{2}{*}{MPPE}
      & DexMachina          & 0.94 & 4.79 & 19.79 & 2.46 & 19.33 & 56.64 & 59.67 & 23.37 \\
      & DexDuet             & \textbf{0.68} & \textbf{1.43} & \textbf{1.31} & \textbf{0.71} & \textbf{0.55} & \textbf{1.40} & \textbf{1.06} & \textbf{1.02} \\
    \bottomrule
  \end{tabular}
\end{table*}

\begin{table*}[!tp]
  \caption{\textbf{Per-demonstration results on the DexMachina suite with XHand hands}, as in \cref{tab:suppl-dexmachina}, with ConTrack~\cite{liang2026contrack} trained with five seeds (\cref{sec:suppl-baselines}).}
  \label{tab:suppl-dexmachina-xhand}
  \centering
  \footnotesize
  \setlength{\tabcolsep}{3pt}
  \begin{tabular}{@{}llcccccccc@{}}
    \toprule
    & Method & Box-200 & Ketchup-100 & Ketchup-300 & Mixer-170 & Mixer-300 & Notebook-300 & Waffleiron-300 & Mean \\
    \midrule
    \multirow{3}{*}{ADD-AUC}
      & DexMachina          & 88.7$\pm$3.3 & 61.6$\pm$43.9 & 33.2$\pm$34.2 & 93.4$\pm$1.3 & 79.4$\pm$3.6 & 14.7$\pm$7.8 & 53.9$\pm$41.3 & 60.7 \\
      & ConTrack            & 50.7$\pm$15.2 & 15.2$\pm$0.2 & 4.0$\pm$0.1 & 41.4$\pm$2.5 & 25.5$\pm$2.1 & 57.4$\pm$13.4 & 20.5$\pm$12.6 & 30.7 \\
      & DexDuet             & \textbf{96.3}$\pm$0.6 & \textbf{94.4}$\pm$1.3 & \textbf{91.4}$\pm$2.3 & \textbf{98.4}$\pm$0.6 & \textbf{89.8}$\pm$17.7 & \textbf{89.3}$\pm$2.3 & \textbf{95.8}$\pm$1.4 & \textbf{93.6} \\
    \midrule
    \multirow{3}{*}{Contact-AUC}
      & DexMachina          & 45.7$\pm$8.4 & 38.9$\pm$35.5 & 17.1$\pm$20.0 & 70.3$\pm$7.7 & 52.8$\pm$3.9 & 7.4$\pm$3.2 & 35.1$\pm$29.9 & 38.2 \\
      & ConTrack            & 13.8$\pm$9.0 & 0.9$\pm$0.4 & 1.8$\pm$0.1 & 23.6$\pm$7.2 & 17.9$\pm$5.0 & 35.6$\pm$12.2 & 6.9$\pm$6.3 & 14.4 \\
      & DexDuet             & \textbf{67.8}$\pm$5.6 & \textbf{75.7}$\pm$2.0 & \textbf{83.7}$\pm$2.4 & \textbf{85.4}$\pm$4.3 & \textbf{79.2}$\pm$20.8 & \textbf{85.6}$\pm$4.4 & \textbf{78.9}$\pm$2.2 & \textbf{79.5} \\
    \midrule
    \multirow{3}{*}{SP-SR}
      & DexMachina          & 99.0 & 59.4 & 19.8 & 99.2 & 99.0 & 0.0 & 59.4 & 62.3 \\
      & ConTrack            & 80.0 & 0.0 & 0.0 & 0.0 & 0.0 & 60.0 & 0.0 & 20.0 \\
      & DexDuet             & \textbf{100.0} & \textbf{100.0} & \textbf{100.0} & \textbf{100.0} & \textbf{100.0} & \textbf{100.0} & \textbf{100.0} & \textbf{100.0} \\
    \midrule
    \multirow{3}{*}{MP-SR}
      & DexMachina          & 99.0 & 59.4 & 19.8 & 99.0 & \textbf{99.0} & 0.0 & 59.4 & 62.2 \\
      & ConTrack            & 0.0 & 0.0 & 0.0 & 0.0 & 0.0 & 0.0 & 0.0 & 0.0 \\
      & DexDuet             & \textbf{100.0} & \textbf{100.0} & \textbf{100.0} & \textbf{100.0} & 80.0 & \textbf{100.0} & \textbf{100.0} & \textbf{97.1} \\
    \midrule
    \multirow{3}{*}{MPPE}
      & DexMachina          & 1.04 & 15.63 & 74.57 & 0.98 & 2.00 & 67.74 & 38.83 & 28.68 \\
      & ConTrack            & 9.59 & 15.45 & 96.47 & 11.15 & 50.66 & 11.08 & 50.47 & 34.98 \\
      & DexDuet             & \textbf{0.67} & \textbf{1.24} & \textbf{1.38} & \textbf{0.58} & \textbf{1.30} & \textbf{1.42} & \textbf{1.07} & \textbf{1.09} \\
    \bottomrule
  \end{tabular}
\end{table*}

\begin{table*}[!tp]
  \caption{\textbf{Per-demonstration results on the DexMachina suite with Schunk hands}, as in \cref{tab:suppl-dexmachina}.}
  \label{tab:suppl-dexmachina-schunk}
  \centering
  \footnotesize
  \setlength{\tabcolsep}{3pt}
  \begin{tabular}{@{}llcccccccc@{}}
    \toprule
    & Method & Box-200 & Ketchup-100 & Ketchup-300 & Mixer-170 & Mixer-300 & Notebook-300 & Waffleiron-300 & Mean \\
    \midrule
    \multirow{2}{*}{ADD-AUC}
      & DexMachina          & 88.3$\pm$1.5 & 92.8$\pm$2.0 & 76.3$\pm$14.0 & 93.9$\pm$2.7 & 46.2$\pm$34.4 & 76.2$\pm$25.6 & 42.3$\pm$38.0 & 73.7 \\
      & DexDuet             & \textbf{94.9}$\pm$0.9 & \textbf{93.0}$\pm$2.7 & \textbf{94.1}$\pm$1.6 & \textbf{99.0}$\pm$0.3 & \textbf{96.2}$\pm$2.3 & \textbf{87.3}$\pm$4.2 & \textbf{94.2}$\pm$1.3 & \textbf{94.1} \\
    \midrule
    \multirow{2}{*}{Contact-AUC}
      & DexMachina          & 45.5$\pm$5.0 & 63.9$\pm$7.3 & 54.8$\pm$11.0 & 70.3$\pm$4.7 & 23.5$\pm$21.1 & 35.7$\pm$15.5 & 22.6$\pm$23.2 & 45.2 \\
      & DexDuet             & \textbf{76.1}$\pm$6.8 & \textbf{74.4}$\pm$4.4 & \textbf{87.7}$\pm$2.5 & \textbf{89.1}$\pm$1.1 & \textbf{89.2}$\pm$3.5 & \textbf{73.0}$\pm$9.0 & \textbf{73.5}$\pm$3.2 & \textbf{80.4} \\
    \midrule
    \multirow{2}{*}{SP-SR}
      & DexMachina          & 99.2 & 99.0 & 79.2 & 99.0 & 59.4 & 79.2 & 39.6 & 79.2 \\
      & DexDuet             & \textbf{100.0} & \textbf{100.0} & \textbf{100.0} & \textbf{100.0} & \textbf{100.0} & \textbf{100.0} & \textbf{100.0} & \textbf{100.0} \\
    \midrule
    \multirow{2}{*}{MP-SR}
      & DexMachina          & 99.0 & 99.0 & 59.4 & 99.0 & 59.4 & 79.2 & 39.6 & 76.4 \\
      & DexDuet             & \textbf{100.0} & \textbf{100.0} & \textbf{100.0} & \textbf{100.0} & \textbf{100.0} & \textbf{100.0} & \textbf{100.0} & \textbf{100.0} \\
    \midrule
    \multirow{2}{*}{MPPE}
      & DexMachina          & 1.10 & 1.42 & 3.98 & 0.95 & 33.38 & 9.37 & 34.66 & 12.12 \\
      & DexDuet             & \textbf{0.77} & \textbf{1.38} & \textbf{1.15} & \textbf{0.54} & \textbf{0.79} & \textbf{1.64} & \textbf{1.16} & \textbf{1.06} \\
    \bottomrule
  \end{tabular}
\end{table*}

\paragraph{SPIDER suite.}
\Cref{tab:suppl-spider} gives the per-demonstration results.
To check our evaluation, we score the trajectories released by SPIDER, using the states that the authors saved.
They reach an ADD-AUC of 44.2, against the 43.6 that SPIDER reports for the same seven demonstrations~\cite{pan2025spider}, and succeed in 15 of 35 runs.
The challenge's evaluation code (\cref{tab:main}) also tests the moving part, so SPIDER's SP-SR falls from 51.4\% to 34.3\% with it.
Most of the difference comes from the lid of Laptop, whose rotation error of about 1.3\,rad on average now counts.
SPIDER fails on Ketchup-1, Ketchup-2 and Notebook with all five seeds.
On these three demonstrations, it tracks the object poorly and its Contact-AUC is close to zero.
It succeeds on Waffleiron with three of the five seeds.

\begin{table*}[!tp]
  \caption{\textbf{Per-demonstration results on the SPIDER suite} (ARCTIC, subject s01, frames 30--230; two Inspire hands, Genesis), in \%. Released: trajectories released with SPIDER~\cite{pan2025spider}, scored from the saved states. All other rows are evaluated with a limit of 400 collision pairs. SPIDER (rerun, open-loop) and DexDuet (five seeds): mean $\pm$ std over seeds, and the mean alone for SPIDER's Contact-AUC. SR: solved runs under SPIDER's own criterion. Bold marks the best value of each metric in each column.}
  \label{tab:suppl-spider}
  \centering
  \footnotesize
  \setlength{\tabcolsep}{3pt}
  \begin{tabular}{@{}llccccccccc@{}}
    \toprule
    & Method & Box & Ketchup-1 & Ketchup-2 & Laptop & Mixer & Notebook & Waffleiron & Mean & SR  \\
    \midrule
    \multirow{3}{*}{ADD-AUC}
      & SPIDER (released)         & 80.8 & 15.0 & 21.1 & 60.3 & 84.3 & 9.4 & 38.3 & 44.2 & 15/35  \\
      & SPIDER (rerun)            & 85.2$\pm$5.2 & 15.8$\pm$3.8 & 18.8$\pm$1.0 & 50.8$\pm$1.5 & 89.9$\pm$0.5 & 9.5$\pm$0.2 & 62.5$\pm$18.6 & 47.5 & 18/35  \\
      & DexDuet (five seeds)      & \textbf{93.2}$\pm$6.0 & \textbf{81.6}$\pm$8.1 & \textbf{88.5}$\pm$1.7 & \textbf{97.6}$\pm$0.6 & \textbf{93.5}$\pm$2.4 & \textbf{72.3}$\pm$27.4 & \textbf{92.5}$\pm$1.8 & \textbf{88.5} & \textbf{34/35}  \\
    \midrule
    \multirow{2}{*}{Contact-AUC}
      & SPIDER (rerun)            & 64.3 & 3.5 & 4.5 & 57.1 & 79.5 & 2.8 & 43.6 & 36.5 & --  \\
      & DexDuet (five seeds)      & \textbf{87.4}$\pm$12.2 & \textbf{70.9}$\pm$6.6 & \textbf{75.7}$\pm$2.2 & \textbf{94.2}$\pm$1.6 & \textbf{85.1}$\pm$4.2 & \textbf{50.7}$\pm$20.6 & \textbf{74.5}$\pm$2.4 & \textbf{76.9} & --  \\
    \bottomrule
  \end{tabular}
\end{table*}

\paragraph{ManipTrans suite.}
\Cref{tab:suppl-maniptrans} gives the per-demonstration ADD-AUC and success rates with the original criteria.
DexDuet meets SPIDER's criterion in at least 99.5\% of the rollouts of every demonstration, whereas ManipTrans succeeds in 34.0\% of its rollouts on Paste-b and 68.2\% on Laptop.

\begin{table*}[p]
  \caption{\textbf{Per-demonstration results on the ManipTrans suite} (OakInk2, two Inspire hands, Isaac Gym, scene with joints), in \%, over 256 rollouts per demonstration. DexDuet: mean over three training seeds. ManipTrans (retrained): retrained with its official protocol, mean over three training seeds (42, 24 and 66), each evaluated with evaluation seed 2026, every rollout started at the window. Mean column: mean $\pm$ std over seeds of the mean over demonstrations. SP-SR$^\ast$ and MP-SR$^\ast$ are the original object criteria of SPIDER and ManipTrans, which test the object root and the object at every frame, unlike SP-SR and MP-SR in \cref{tab:main}. Bold marks the best value of each metric in each column.}
  \label{tab:suppl-maniptrans}
  \centering
  \footnotesize
  \setlength{\tabcolsep}{3pt}
  \begin{tabular}{@{}llccccccccccc@{}}
    \toprule
    & Method & Gate-a & Gate-b & Gate-c & Gate-d & Gate-e & Gate-f & Gate-g & Laptop & Paste-a & Paste-b & Mean \\
    \midrule
    \multirow{2}{*}{ADD-AUC}
      & ManipTrans (retrained) & 98.6 & \textbf{100.0} & \textbf{99.0} & \textbf{99.9} & \textbf{100.0} & 77.2 & 93.8 & 71.7 & 93.1 & 37.2 & 87.0$\pm$4.1 \\
      & DexDuet                & \textbf{99.5} & 95.5 & \textbf{99.0} & 99.5 & 99.8 & \textbf{99.3} & \textbf{95.5} & \textbf{99.9} & \textbf{97.4} & \textbf{98.6} & \textbf{98.4}$\pm$0.4 \\
    \midrule
    \multirow{2}{*}{SP-SR$^\ast$}
      & ManipTrans (retrained) & 98.7 & \textbf{100.0} & 98.8 & 99.9 & \textbf{100.0} & 86.2 & 96.1 & 68.2 & 90.8 & 34.0 & 87.3$\pm$4.2 \\
      & DexDuet                & \textbf{100.0} & \textbf{100.0} & \textbf{100.0} & \textbf{100.0} & \textbf{100.0} & \textbf{100.0} & \textbf{100.0} & \textbf{100.0} & \textbf{99.5} & \textbf{99.9} & \textbf{99.9}$\pm$0.1 \\
    \midrule
    \multirow{2}{*}{MP-SR$^\ast$}
      & ManipTrans (retrained) & 98.0 & \textbf{99.9} & 98.4 & 99.9 & 99.9 & 63.3 & 91.7 & 58.6 & \textbf{90.2} & 33.5 & 83.3$\pm$4.0 \\
      & DexDuet                & \textbf{99.7} & 92.6 & \textbf{99.3} & \textbf{100.0} & \textbf{100.0} & \textbf{100.0} & \textbf{99.6} & \textbf{100.0} & 49.2 & \textbf{81.0} & \textbf{92.1}$\pm$5.1 \\
    \bottomrule
  \end{tabular}
\end{table*}

\paragraph{V2D challenge.}
DexDuet exceeds CHORD's leaderboard entry and both retrained CHORD runs on all four tracking metrics (\cref{tab:suppl-v2d}).
The two interaction rewards add 6.0 points of Contact-AUC at a similar ADD-AUC (\cref{tab:suppl-v2d}).
DexDuet reaches a higher ADD-AUC than CHORD (PPO) on 28 of the 30 episodes (\cref{tab:suppl-v2d-episodes}).

\begin{table*}[!tp]
  \caption{\textbf{V2D challenge} (30 episodes, G1 + Dex3, official evaluator), in \%, and MPPE in cm. Steps: environment steps per episode in billions (range and mean). For the runs that we train, each value is the mean of four runs of the evaluator, and Contact-AUC is averaged over 29 episodes (\cref{sec:suppl-contact-points}). The second block, except DexDuet without interaction rewards, gives the V2D rows of \cref{tab:main}. Bold marks the best value in each column.}
  \label{tab:suppl-v2d}
  \centering
  \footnotesize
  \begin{tabular}{@{}llccccc@{}}
    \toprule
    Method & Steps & ADD-AUC$\uparrow$ & SP-SR$\uparrow$ & MP-SR$\uparrow$ & MPPE$\downarrow$ & Contact-AUC$\uparrow$ \\
    \midrule
    CHORD~\cite{zhu2026chord}, leaderboard entry  & not published           & 51.0 & 70.0 & 20.0 & 17.74 & -- \\
    \midrule
    CHORD (PPO, retrained)                        & 1.97                    & 39.8 & 53.3 & 15.8 & 14.85 & 16.0 \\
    CHORD (FlashSAC, retrained)                        & 1.97                    & 32.0 & 35.0 & 12.5 & 37.77 & 13.0 \\
    DexDuet without interaction rewards           & 1.70--1.97 (mean 1.92)  & 76.2 & 83.3 & 71.7 & \textbf{11.47} & 66.6 \\
    DexDuet                                       & 1.70--1.97 (mean 1.91)  & \textbf{77.4} & \textbf{85.0} & \textbf{75.8} & 12.19 & \textbf{72.6} \\
    \bottomrule
  \end{tabular}
\end{table*}

\begin{table*}[!tp]
  \caption{\textbf{Per-episode results on the V2D challenge} (G1 + Dex3, official evaluator), in \%. Each value is the mean of four runs of the evaluator, and Contact-AUC is computed as in \cref{sec:suppl-contact-points}. PPO and FlashSAC: CHORD trained with the two recipes; w/o IR: DexDuet without interaction rewards. In episode 29 the human never holds the object, so it has no Contact-AUC. Bold marks the best value of each metric in each row. The last row gives the means of \cref{tab:suppl-v2d}.}
  \label{tab:suppl-v2d-episodes}
  \centering
  \footnotesize
  \setlength{\tabcolsep}{4pt}
  \begin{tabular}{@{}rlcccccccc@{}}
    \toprule
    & & \multicolumn{4}{c}{ADD-AUC$\uparrow$} & \multicolumn{4}{c}{Contact-AUC$\uparrow$} \\
    \cmidrule(lr){3-6} \cmidrule(l){7-10}
    Ep. & Object & PPO & FlashSAC & w/o IR & DexDuet & PPO & FlashSAC & w/o IR & DexDuet \\
    \midrule
    0 & bar stool, tip up & 52.8 & 49.8 & \textbf{99.8} & 99.7 & 33.8 & 17.3 & 93.7 & \textbf{97.3} \\
    1 & vacuum & 18.6 & \textbf{35.4} & 19.2 & 19.6 & 3.6 & \textbf{12.4} & 5.9 & 7.3 \\
    2 & Easter basket & 44.4 & 40.9 & \textbf{91.5} & 90.2 & 13.2 & 20.1 & \textbf{94.4} & 92.2 \\
    3 & mug & 52.5 & 7.6 & 97.5 & \textbf{98.4} & 25.3 & 0.0 & 72.7 & \textbf{98.5} \\
    4 & baseball bat & 13.7 & \textbf{16.1} & 13.5 & 13.2 & \textbf{0.0} & \textbf{0.0} & \textbf{0.0} & \textbf{0.0} \\
    5 & box & 51.7 & 16.6 & \textbf{94.3} & 93.0 & 29.8 & 0.0 & \textbf{59.8} & 59.6 \\
    6 & monitor & 58.6 & 72.0 & 88.8 & \textbf{91.0} & 31.2 & 78.9 & \textbf{98.8} & 97.9 \\
    7 & iron & 38.9 & 22.8 & \textbf{100.0} & 99.9 & 6.5 & 0.0 & 99.4 & \textbf{99.5} \\
    8 & wooden chair & 70.1 & 74.2 & \textbf{96.6} & \textbf{96.6} & 41.0 & 55.8 & 82.5 & \textbf{92.5} \\
    9 & bowl & 23.0 & 13.6 & 45.1 & \textbf{82.1} & 0.3 & 0.0 & 24.0 & \textbf{95.1} \\
    10 & desk & 26.9 & 25.6 & \textbf{75.2} & 74.8 & 12.9 & 19.5 & \textbf{48.0} & 32.3 \\
    11 & book & 35.2 & 8.5 & 86.8 & \textbf{93.2} & 11.4 & 0.0 & 44.4 & \textbf{94.9} \\
    12 & platform & 45.3 & 62.1 & \textbf{96.3} & 96.0 & 25.2 & 28.1 & 84.7 & \textbf{89.5} \\
    13 & snack box & 49.3 & 42.1 & \textbf{81.9} & 56.8 & 19.9 & 25.0 & \textbf{82.1} & 31.8 \\
    14 & round table & 42.0 & 37.1 & 66.9 & \textbf{73.9} & 38.1 & 17.0 & 62.7 & \textbf{75.8} \\
    15 & wet-floor sign & 46.6 & 23.8 & \textbf{93.6} & 77.8 & 9.1 & 0.0 & \textbf{98.0} & 72.3 \\
    16 & spray bottle & 55.0 & 45.4 & \textbf{96.0} & 94.2 & 9.3 & 8.7 & 96.2 & \textbf{98.7} \\
    17 & traffic cone & 40.3 & 28.4 & 92.6 & \textbf{93.6} & 7.9 & 0.0 & 95.4 & \textbf{95.9} \\
    18 & trash can & 45.7 & 58.2 & \textbf{98.2} & 97.2 & 25.0 & 25.7 & 94.3 & \textbf{94.9} \\
    19 & cane & 12.3 & 14.1 & 15.1 & \textbf{16.2} & 0.0 & 0.0 & 4.6 & \textbf{5.7} \\
    20 & rice cooker & 45.5 & 30.4 & \textbf{97.1} & 92.6 & 37.9 & 23.8 & \textbf{92.3} & 90.6 \\
    21 & bar stool, pull & 28.3 & 13.8 & \textbf{33.5} & 25.3 & 6.4 & 0.0 & \textbf{22.8} & 11.3 \\
    22 & step ladder & 53.6 & 29.3 & \textbf{95.1} & 92.2 & 19.7 & 0.0 & \textbf{97.4} & 96.2 \\
    23 & toy stroller & 47.0 & 31.7 & \textbf{98.4} & 97.7 & 20.1 & 4.8 & 96.7 & \textbf{97.1} \\
    24 & watering can & 33.7 & 45.5 & 62.5 & \textbf{99.8} & 5.0 & 26.7 & 19.6 & \textbf{99.9} \\
    25 & water bottle & 13.0 & 11.8 & 26.7 & \textbf{43.4} & 3.1 & 0.0 & 30.7 & \textbf{45.7} \\
    26 & skateboard & 49.3 & 47.5 & \textbf{99.5} & 99.3 & 21.6 & 11.7 & \textbf{95.7} & 93.3 \\
    27 & basket & 38.8 & 27.0 & 95.8 & \textbf{99.4} & 4.1 & 0.0 & 98.2 & \textbf{99.9} \\
    28 & squeegee & 17.2 & 13.5 & 45.9 & \textbf{48.1} & 3.3 & 2.7 & 37.5 & \textbf{40.2} \\
    29 & rolling luggage & 44.0 & 16.1 & \textbf{82.9} & 66.4 & -- & -- & -- & -- \\
    \midrule
    & Mean & 39.8 & 32.0 & 76.2 & \textbf{77.4} & 16.0 & 13.0 & 66.6 & \textbf{72.6} \\
    \bottomrule
  \end{tabular}
\end{table*}

\subsection{Holding After the Last Frame}
\label{sec:suppl-hold}

On the DexMachina suite, we re-simulate each policy from the simulator states of its own rollout and let it keep acting for 3\,s after the last frame, with every reference frozen at the last frame.
A rollout holds the object if the object stays within 5\,cm of its final pose for the whole 3\,s, and drops it if the object ends the 3\,s more than 10\,cm below its final pose or off the table.
We run it with all four hands on the checkpoints of \cref{tab:main}.
DexDuet holds the object in more runs and drops it in fewer runs than DexMachina with every hand (\cref{tab:suppl-hold-hands}).

\begin{table*}[!tp]
  \caption{\textbf{Holding after the last frame} with four hands (DexMachina suite), in \% of rollouts, mean $\pm$ std over the five training seeds. Off track: number of runs that end more than 3\,cm from the final pose, of 35 per hand and 140 in all.}
  \label{tab:suppl-hold-hands}
  \centering
  \footnotesize
  \setlength{\tabcolsep}{3pt}
  \newcommand{\stdv}[1]{{\tiny$\pm$#1}}
  \begin{tabular}{@{}llccc@{}}
    \toprule
    Hand & Method & Held$\uparrow$ & Dropped$\downarrow$ & Off track$\downarrow$ \\
    \midrule
    \multirow{2}{*}{Inspire} & DexMachina & 17.0\stdv{11.8} & 71.7\stdv{14.2} & 18 \\
                             & DexDuet    & \textbf{37.1}\stdv{16.3} & \textbf{51.4}\stdv{7.8} & \textbf{5} \\
    \addlinespace[2pt]
    \multirow{2}{*}{Allegro} & DexMachina & 14.1\stdv{10.0} & 80.0\stdv{7.7} & 16 \\
                             & DexDuet    & \textbf{22.9}\stdv{12.8} & \textbf{62.9}\stdv{16.3} & \textbf{5} \\
    \addlinespace[2pt]
    \multirow{2}{*}{XHand}   & DexMachina & 5.7\stdv{7.7} & 71.7\stdv{17.3} & 18 \\
                             & DexDuet    & \textbf{25.7}\stdv{15.6} & \textbf{54.3}\stdv{18.6} & \textbf{1} \\
    \addlinespace[2pt]
    \multirow{2}{*}{Schunk}  & DexMachina & 25.5\stdv{11.8} & 68.9\stdv{15.5} & 10 \\
                             & DexDuet    & \textbf{42.9}\stdv{10.1} & \textbf{51.4}\stdv{16.3} & \textbf{0} \\
    \midrule
    \multirow{2}{*}{All}     & DexMachina & 15.6\stdv{3.2} & 73.1\stdv{6.9} & 62 \\
                             & DexDuet    & \textbf{32.1}\stdv{7.1} & \textbf{55.0}\stdv{6.0} & \textbf{11} \\
    \bottomrule
  \end{tabular}
\end{table*}

\subsection{Ablations}
\label{sec:suppl-ablation}

Each row of \cref{tab:ablation} removes or replaces components of DexDuet and keeps everything else, including the reward weights of \cref{tab:suppl-reward} and the settings of \cref{tab:suppl-clips}.

\begin{table*}[!tp]
  \caption{\textbf{Ablations, all metrics} (DexMachina suite, Inspire, seeds 15, 42 and 113), in \%, and MPPE in cm. Means over the seven demonstrations. Worse / better: paired runs (same demonstration and seed) in which the row is below / above DexDuet (a), out of 21. Held: share of rollouts that hold the object after the last frame (\cref{sec:suppl-hold}). Bold marks the best value in each column.}
  \label{tab:suppl-ablation-all}
  \centering
  \footnotesize
  \setlength{\tabcolsep}{4pt}
  \begin{tabular}{@{}llccccccccc@{}}
    \toprule
    & & & & & & & \multicolumn{2}{c}{Worse / better} & \\
    \cmidrule(lr){8-9}
    Row & Change & ADD-AUC$\uparrow$ & Contact-AUC$\uparrow$ & SP-SR$\uparrow$ & MP-SR$\uparrow$ & MPPE$\downarrow$ & ADD & Contact & Updates$\downarrow$ & Held$\uparrow$ \\
    \midrule
    (a) & DexDuet                                                  & \textbf{89.7} & \textbf{80.5} & 95.2  & 95.2  & \textbf{1.45}  & --      & --     & \textbf{5,524} & \textbf{47.6} \\
    (b) & $\lambda_{\text{hg}}=0$                                  & 87.5 & 71.5 & \textbf{100.0} & \textbf{100.0} & 1.56  & 16 / 5  & 17 / 4 & 6,721 & 28.6 \\
    (c) & $\lambda_{\text{hoc}}=0$                                 & 86.4 & 65.4 & 85.7  & 85.7  & 3.01  & 8 / 13  & 18 / 3 & 5,836 & 33.3 \\
    (d) & $\lambda_{\text{hg}}=\lambda_{\text{hoc}}=0$             & 76.7 & 49.0 & 71.4  & 71.4  & 7.96  & 12 / 9  & 21 / 0 & 6,745 & 38.1 \\
    (e) & unified policy                                           & 74.1 & 62.1 & \textbf{100.0} & 90.5  & 2.58  & 20 / 1  & 19 / 2 & 6,255 & 14.3 \\
    (f) & (d) with single-level VOC                                          & 79.4 & 47.6 & 76.2  & 76.2  & 15.92 & 8 / 13  & 21 / 0 & 6,845 & 28.6 \\
    (g) & single-level VOC, no A-RSI                                  & 88.4 & 74.2 & 95.2  & 95.2  & 1.79  & 10 / 11 & 16 / 5 & 6,608 & 23.8 \\
    \bottomrule
  \end{tabular}
\end{table*}

\begin{table*}[!tp]
  \caption{\textbf{Ablations per demonstration} (DexMachina suite, Inspire), in \%: mean $\pm$ std over training seeds 15, 42 and 113. Updates: mean number of PPO updates until training stops, with the caps of \cref{tab:suppl-clips}. Bold marks the best value of each metric in each column.}
  \label{tab:suppl-ablation}
  \centering
  \footnotesize
  \setlength{\tabcolsep}{3pt}
  \begin{tabular}{@{}llcccccccc@{}}
    \toprule
    & Row & Box-200 & Ketchup-100 & Ketchup-300 & Mixer-170 & Mixer-300 & Notebook-300 & Waffleiron-300 & Mean \\
    \midrule
    \multirow{7}{*}{ADD-AUC}
      & (a) & 94.9$\pm$1.4 & 79.6$\pm$15.5 & 87.2$\pm$2.9 & 95.6$\pm$1.8 & 93.1$\pm$3.0 & 86.7$\pm$4.3 & 91.1$\pm$2.0 & \textbf{89.7} \\
      & (b) & 95.1$\pm$1.0 & 87.9$\pm$1.9 & 85.0$\pm$4.3 & 90.8$\pm$1.3 & 89.0$\pm$4.7 & 74.0$\pm$1.7 & 90.9$\pm$1.6 & 87.5 \\
      & (c) & 96.4$\pm$1.6 & 46.1$\pm$18.3 & \textbf{90.7}$\pm$1.6 & \textbf{98.7}$\pm$0.3 & 93.5$\pm$6.7 & 87.8$\pm$5.1 & 91.7$\pm$1.6 & 86.4 \\
      & (d) & 97.4$\pm$1.0 & 69.6$\pm$19.8 & 70.9$\pm$25.9 & 98.2$\pm$0.8 & 82.9$\pm$3.1 & 68.8$\pm$42.4 & 49.4$\pm$40.8 & 76.7 \\
      & (e) & 85.1$\pm$2.6 & 80.2$\pm$9.2 & 66.0$\pm$2.9 & 83.8$\pm$0.7 & 63.8$\pm$8.2 & 74.4$\pm$3.9 & 65.5$\pm$8.3 & 74.1 \\
      & (f) & \textbf{97.6}$\pm$0.6 & 60.3$\pm$32.7 & 22.9$\pm$14.9 & 97.9$\pm$0.9 & \textbf{94.6}$\pm$3.2 & \textbf{90.0}$\pm$1.9 & \textbf{92.4}$\pm$1.2 & 79.4 \\
      & (g) & 91.8$\pm$10.4 & \textbf{89.7}$\pm$2.6 & 74.5$\pm$11.1 & 97.3$\pm$0.3 & 91.8$\pm$5.1 & 85.1$\pm$4.6 & 88.4$\pm$2.0 & 88.4 \\
    \midrule
    \multirow{7}{*}{Contact-AUC}
      & (a) & \textbf{92.2}$\pm$2.4 & 63.3$\pm$24.3 & \textbf{79.4}$\pm$5.8 & \textbf{86.7}$\pm$4.3 & \textbf{88.6}$\pm$2.1 & \textbf{80.5}$\pm$1.8 & \textbf{73.1}$\pm$1.0 & \textbf{80.5} \\
      & (b) & 85.3$\pm$4.3 & 78.9$\pm$4.5 & 66.4$\pm$4.6 & 80.4$\pm$4.4 & 73.8$\pm$4.7 & 46.8$\pm$4.0 & 68.9$\pm$3.4 & 71.5 \\
      & (c) & 86.1$\pm$8.3 & 12.0$\pm$12.4 & 71.5$\pm$4.3 & \textbf{86.7}$\pm$1.5 & 77.9$\pm$10.6 & 66.3$\pm$3.1 & 57.2$\pm$7.8 & 65.4 \\
      & (d) & 74.1$\pm$3.3 & 31.0$\pm$37.3 & 38.7$\pm$22.5 & 74.5$\pm$4.0 & 51.3$\pm$1.3 & 45.8$\pm$33.9 & 27.7$\pm$23.2 & 49.0 \\
      & (e) & 84.7$\pm$7.6 & 70.7$\pm$11.9 & 39.5$\pm$7.3 & 75.4$\pm$3.9 & 52.1$\pm$17.9 & 67.0$\pm$3.6 & 45.3$\pm$11.3 & 62.1 \\
      & (f) & 70.3$\pm$4.2 & 15.0$\pm$16.4 & 5.0$\pm$4.4 & 70.8$\pm$3.6 & 63.3$\pm$2.8 & 57.3$\pm$6.1 & 51.4$\pm$3.2 & 47.6 \\
      & (g) & 80.9$\pm$22.0 & \textbf{81.6}$\pm$2.8 & 53.3$\pm$4.4 & 85.7$\pm$3.0 & 80.9$\pm$5.9 & 67.8$\pm$16.5 & 69.2$\pm$1.1 & 74.2 \\
    \midrule
    \multirow{7}{*}{Updates}
      & (a) & 4,530 & 4,987 & \textbf{4,757} & 4,803 & 6,420 & 8,623 & \textbf{4,550} & \textbf{5,524} \\
      & (b) & 4,510 & 4,963 & 7,367 & 4,830 & 8,400 & 11,500 & 5,477 & 6,721 \\
      & (c) & 4,600 & 5,000 & 6,947 & 4,420 & 6,563 & \textbf{7,513} & 5,810 & 5,836 \\
      & (d) & \textbf{4,403} & 5,000 & 7,493 & 4,570 & 8,610 & 8,803 & 8,333 & 6,745 \\
      & (e) & 4,430 & \textbf{4,597} & 10,377 & \textbf{4,377} & 6,120 & 9,113 & 4,773 & 6,255 \\
      & (f) & 4,627 & 5,000 & 13,000 & 4,573 & \textbf{5,207} & 9,443 & 6,063 & 6,845 \\
      & (g) & 4,673 & 4,980 & 13,000 & 4,907 & 6,190 & 7,613 & 4,893 & 6,608 \\
    \bottomrule
  \end{tabular}
\end{table*}

\paragraph{Variants.}
Rows (b) and (c) set $\lambda_{\text{hg}}$ or $\lambda_{\text{hoc}}$ to zero, and row (d) sets both to zero.
Row (e) replaces the two per-hand policies by the unified policy of DexMachina~\cite{mandi2026dexmachina}, an MLP with hidden layers of 512, 512, 256 and 128 units shared by the actor and the critic.
It maps the observation of both hands to their concatenated 24-dimensional action, with one value head and one PPO ratio for the joint action.
Rows (f) and (g) replace A-RSI by DexMachina's single-level VOC curriculum (\cref{sec:suppl-curriculum}), starting from row (d) and from DexDuet, respectively.

\paragraph{Training and evaluation.}
Every row is trained with seeds 15, 42 and 113 under the stopping rule of \cref{sec:suppl-impl} and the update caps of \cref{tab:suppl-clips}.
\Cref{tab:suppl-ablation-all} adds the other metrics, the comparison with row (a) paired by demonstration and seed, and the share of rollouts that hold the object after the last frame, and \cref{tab:suppl-ablation} gives each demonstration.

\paragraph{Interaction rewards.}
The interaction rewards act mainly on the contacts.
Removing the hand graph reward or the hand-object contact reward lowers Contact-AUC by 9.0 and 15.1 points, against 2.2 and 3.3 points of ADD-AUC.
The two rewards matter most on different demonstrations.
Without the hand graph reward, Notebook-300 loses the most, 12.7 points of ADD-AUC and 33.7 points of Contact-AUC, and without the hand-object contact reward, all three seeds track the bottle poorly on Ketchup-100.
Removing both (d) makes training less reliable: six of its 21 runs fail SP-SR, two each on Ketchup-100 and Waffleiron-300 and one each on Ketchup-300 and Notebook-300, against one run of DexDuet (\cref{tab:suppl-ablation-all}).
On the five demonstrations on which (f), the variant without the interaction rewards and A-RSI, tracks the object better than DexDuet, by 1.3 to 3.3 points of ADD-AUC, DexDuet preserves the human's contacts better by 16 to 25 points of Contact-AUC.

\paragraph{Per-hand policy.}
The unified policy (e) is below DexDuet in both metrics in every paired run on six of the seven demonstrations.

\paragraph{A-RSI.}
A-RSI matters most where the single assistance level stalls.
On Ketchup-300, all six runs of (f) and (g) reach the update cap of 13,000 and end with a lower ADD-AUC and Contact-AUC than their counterparts with A-RSI, (d) and (a).
With the interaction rewards, row (a) reaches a higher Contact-AUC than (g) on six of the seven demonstrations.

\subsection{Real-World Replays}
\label{sec:suppl-real}

\begin{figure*}[!tp]
  \centering
  \includegraphics[width=0.78\linewidth]{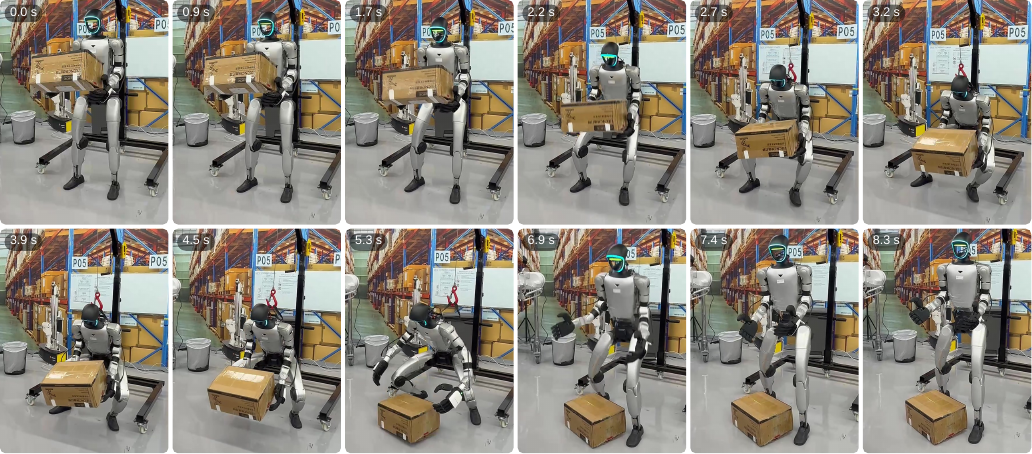}\\[2pt]
  {\small (a) Box placement}\\[6pt]
  \includegraphics[width=0.78\linewidth]{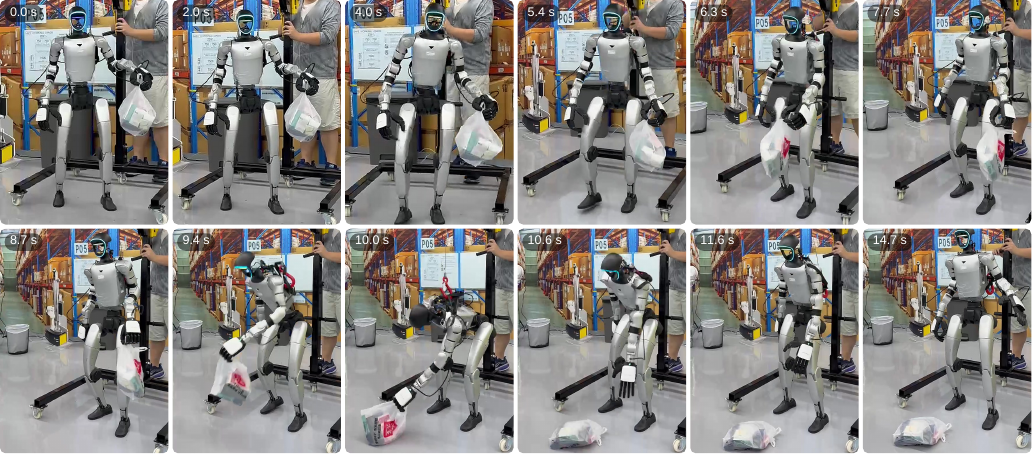}\\[2pt]
  {\small (b) Bag placement}\\[6pt]
  \includegraphics[width=0.78\linewidth]{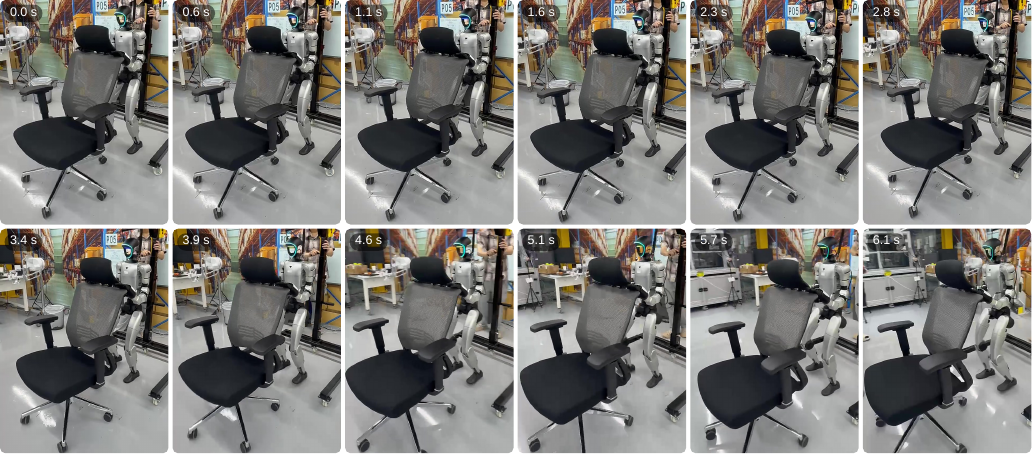}\\[2pt]
  {\small (c) Chair pushing}
  \caption{\textbf{Real-world replays on the Unitree G1} with Inspire hands. (a) The robot lowers a box from chest height, places it on the floor and stands up. (b) It carries a bag and puts it down on the floor. (c) It pushes an office chair while walking.}
  \label{fig:suppl-real-g1}
\end{figure*}

\begin{figure*}[!tp]
  \centering
  \includegraphics[width=\linewidth]{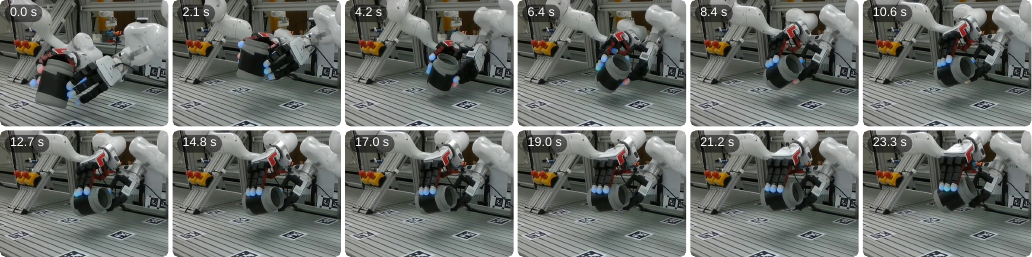}\\[2pt]
  {\small (a) Bimanual object reorientation}\\[6pt]
  \includegraphics[width=\linewidth]{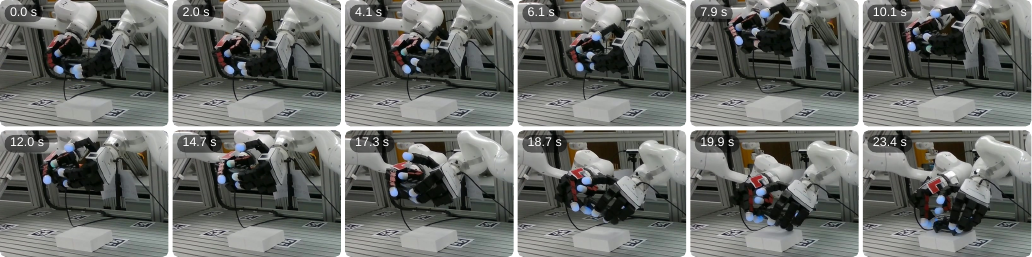}\\[2pt]
  {\small (b) Bimanual object placement}
  \caption{\textbf{Real-world replays on two Franka arms} with Allegro hands. (a) Both hands reorient a bottle that they hold above the table. (b) Both hands place a gamepad on a block.}
  \label{fig:suppl-real-franka}
\end{figure*}

\Cref{fig:suppl-real-g1,fig:suppl-real-franka} show the replays of \cref{sec:real} frame by frame.
On the Unitree G1 with Inspire hands, the robot lowers a box and places it on the floor, carries a bag and puts it down, and pushes an office chair while walking.
On the two Franka arms with Allegro hands, both hands hold a bottle above the table and reorient it, and place a gamepad, held by both hands, on a block.

\section{Limitations}
\label{sec:suppl-limitations}
Like DexMachina and ManipTrans, DexDuet trains one policy per demonstration, which takes about a day on one GPU, and it does not generalize to new objects or motions.
The policy observes the object state of the simulator, and the demonstration must provide accurate hand and object motion.
On the real robots, we replay simulated rollouts open-loop (\cref{sec:real}), and closed-loop deployment with an estimated object state is left to future work.

\end{document}